\documentclass{article}

\PassOptionsToPackage{numbers,sort&compress}{natbib}
 \usepackage[preprint]{neurips_2026}

\usepackage[utf8]{inputenc} 
\usepackage[T1]{fontenc}    
\usepackage{hyperref}       
\usepackage{url}            
\usepackage{booktabs}       
\usepackage{amsfonts}       
\usepackage{nicefrac}       
\usepackage{microtype}      
\usepackage{xcolor}         
\usepackage{amsmath}
\usepackage{graphicx}
\usepackage{xspace}
\usepackage{wrapfig}
\usepackage{multirow}
\usepackage{colortbl}
\usepackage{xcolor}
\usepackage{algorithm}
\usepackage{algpseudocode}
\usepackage{float}

\definecolor{gold}{HTML}{FFD700}
\definecolor{silver}{HTML}{E8C9FF}
\definecolor{bronze}{HTML}{FFE0B2}

\newcommand{\first}[1]{\cellcolor{gold}\textbf{#1}}
\newcommand{\second}[1]{\cellcolor{silver}{#1}}

\newcommand{\alg}{SplatShot\xspace}

\title{SplatShot: 3D Face Avatar Generation from a Single Unconstrained Photo}

\author{%
  Hao Liang \\
  Rice University\\
  Houston, TX 77005 \\
  \texttt{hl106@rice.edu} \\
  \And
  Zhixuan Ge \\
  Rice University\\
  Houston, TX 77005 \\
  \texttt{zg33@rice.edu} \\
  \AND
  Soumendu Majee \\
  Samsung Research America \\
  Plano, TX 75023 \\
  \texttt{email} \\
  \And
  Joanni Li \\
  Rice University\\
  Houston, TX 77005 \\
  \texttt{jl561@rice.edu} \\
  \And
  Ashok Veeraraghavan \\
  Rice University\\
  Houston, TX 77005 \\
  \texttt{vashok@rice.edu} \\
  \And
  Guha Balakrishnan \\
  Rice University\\
  Houston, TX 77005 \\
  \texttt{guha@rice.edu} \\
}

\begin{document}

\maketitle

\begin{figure}[h]
    \centering
    \includegraphics[width=\linewidth, trim={20 10 10 10},  
    clip]{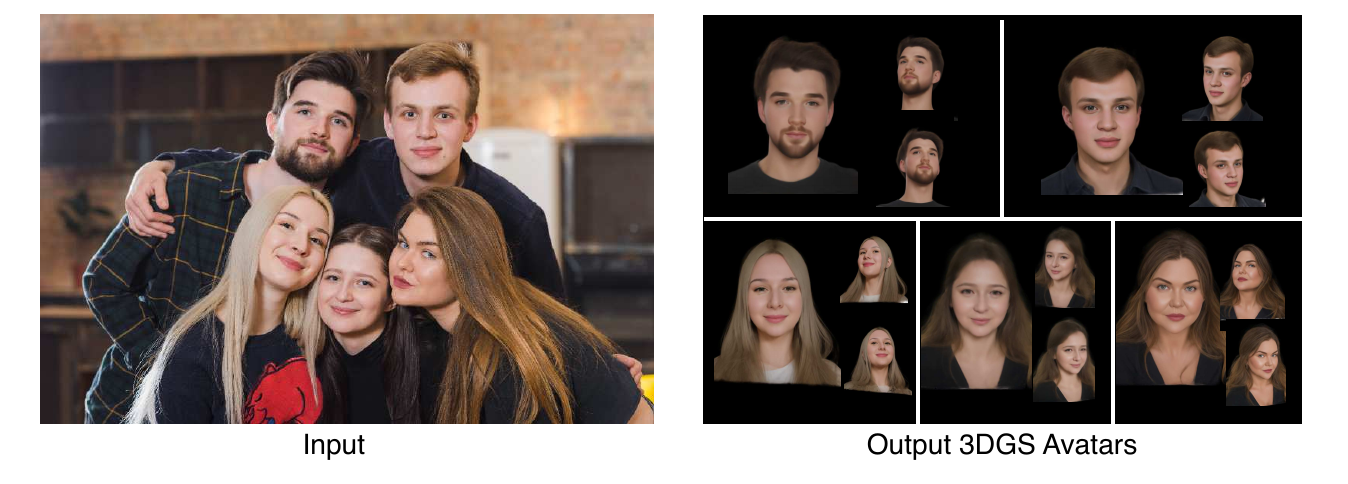}
    \caption{\textbf{From a casual group photo to individual 3D face avatars.} Given a single unconstrained photograph (left), SplatShot produces an explicit 3D Gaussian Splatting (3DGS)~\cite{kerbl20233d} face avatar for each individual that can be rendered from novel viewpoints (right). }
    \label{fig:teaser}
\end{figure}

\begin{abstract}
Reconstructing a photorealistic 3D face avatar from a single unconstrained photograph is challenging: feed-forward 3D Gaussian Splatting (3DGS) models degrade on out-of-distribution inputs, while pretrained diffusion models produce high-fidelity images but lack multi-view consistency. We observe that these paradigms are fundamentally complementary: explicit 3D representations guarantee geometric consistency, whereas 2D diffusion priors ensure photorealism. Building on this, we propose \textbf{SplatShot}, a training-free framework that couples these representations directly within the denoising process. Given a base 3DGS face model and a single reference image, we jointly denoise all target views using a per-step 3D feedback loop. At each timestep, we predict clean images from the noisy latents, refit the 3DGS to these multi-view predictions, and back-propagate the photometric discrepancy between the 3DGS re-renderings and 2D predictions into the noise estimate. This steers the sampling trajectory toward strictly 3D-coherent, identity-faithful outputs. Experiments on diverse in-the-wild images demonstrate that \textbf{SplatShot} produces 3D avatars with superior identity preservation, photorealism, and multi-view consistency. Code is available at \href{https://github.com/hliang2/SplatShot}{\texttt{https://github.com/hliang2/SplatShot}}.
\end{abstract}    
\section{Introduction}
\label{sec:intro}

3D face avatar generation has broad applications in telepresence, augmented and virtual reality, gaming, and digital content creation. The ultimate, yet most challenging, objective is taking a single ``in-the-wild'' portrait, captured under arbitrary lighting and pose, and producing a complete 3D representation that is photorealistic from all viewpoints while faithfully preserving the subject's identity. Despite significant progress, this remains an open problem: existing methods struggle to simultaneously maintain strict geometric consistency, preserve identity, and generalize to arbitrary, out-of-distribution images.

Current approaches fall into three categories, each facing distinct trade-offs. First, explicit 3D representations (e.g., NeRF~\cite{mildenhall2021nerf}, 3DGS~\cite{kerbl20233d}) achieve high quality via per-subject optimization but require dense multi-view captures. Feed-forward 3D variants(e.g., GAGAvatar~\cite{chu2024generalizable}, FastAvatar~\cite{liang2025fastavatar}) alleviate this capture requirement but generalize poorly to in-the-wild inputs due to domain-specific training. Second, 2D generative models trained with 3D awareness(e.g., EG3D~\cite{chan2022efficient}) generate highly photorealistic faces but sample views independently, resulting in severe multi-view inconsistency. Even multi-view trained diffusion models~\cite{gerogiannis2025arc2avatar,xue2024human} suffer from over-smoothing and domain sensitivity. Finally, Score Distillation Sampling (SDS)~\cite{poole2022dreamfusion} and its variants use 2D priors to guide 3D generation but notoriously exhibit mode-seeking behavior. This yields over-saturated, low-diversity outputs that fail to balance identity fidelity with geometric plausibility.

We start from a core observation: explicit 3D models and 2D diffusion models have complementary weaknesses. An explicit 3DGS representation naturally enforces multi-view consistency, but its rendering quality is bottlenecked by the input views used to fit it. Conversely, a pretrained diffusion model provides strong priors for high-fidelity image generation but lacks a mechanism to coordinate outputs across disparate views. By forcing these representations to inform each other during generation, we achieve results that are simultaneously high-quality and strictly 3D-consistent.

This insight forms the basis of our framework, \alg. We assume a single given input photograph defining the target identity. Starting from an existing 3DGS face model (of a prior identity) to provide a structured initialization, we initiate an iterative diffusion denoising process with this model in a feedback loop. At each denoising timestep, we: (1)~predict the clean images $\hat{\mathbf{x}}_0$ from the current noisy latents, (2)~refit the 3DGS model to these multi-view predictions, (3)~re-render from the updated 3DGS, and (4)~back-propagate the photometric discrepancy between the re-renderings and predictions into the noise estimate. At early denoising steps (high noise levels), we observe that $\hat{\mathbf{x}}_0$ is heavily corrupted and unsuitable for supervising 3DGS fitting. We address this via a noise mixture mechanism, producing a geometry-anchored estimate robust enough for guidance even at high noise levels. As denoising progresses, the diffusion model drives realism while the 3DGS model enforces tight geometric agreement, ultimately yielding a photorealistic and 3D-coherent Gaussian avatar.

Crucially, our framework is training-free, requires no fine-tuning of any model component, and works out-of-the-box with off-the-shelf pretrained models. Extensive evaluations on a diverse set of in-the-wild images demonstrate that \alg outperforms representative baselines across multiple prior categories. To the best of our knowledge, we are the first to demonstrate high-fidelity, unconstrained 3D Gaussian face avatar generation from an arbitrary single in-the-wild image without requiring lengthy subject-specific optimization. By producing 3DGS representations with superior identity preservation, photorealism, and strict multi-view consistency, \alg establishes a new state-of-the-art for practical single-image 3D avatar creation.

\section{Related Work}
\label{sec:related}
We discuss the most representative related work below and provide an expanded list in Appendix.

\subsection{Single-Image 3D Face and Head Reconstruction}
Reconstructing 3D faces from images has evolved from parametric morphable models~\cite{blanz2003face}, which offer semantic control but lack fine details, to neural rendering approaches like NeRF~\cite{mildenhall2021nerf} and 3D Gaussian Splatting (3DGS)~\cite{kerbl20233d}. While the latter achieve high photorealism, per-subject optimization typically requires dense, calibrated multi-view captures. To relax this requirement, feed-forward methods predict 3DGS avatars directly from single images \cite{ki2024learning}. However, their quality is heavily bounded by training distributions; models trained on curated datasets degrade on in-the-wild inputs, and multi-view diffusion often yields over-smoothed outputs, yet lacking geometry consistency. \alg sidesteps these domain gaps by coupling an explicit 3D representation with a broadly pretrained 2D diffusion model strictly at inference time, eliminating the need for task-specific training.

\subsection{Coupling 2D Models with 3D Representations}

2D generative models have achieved remarkable photorealism for face synthesis, from VAEs~\cite{kingma2013auto}, GANs~\cite{karras2019style} to diffusion models~\cite{rombach2022high}. Extending these to 3D, methods such as EG3D~\cite{chan2022efficient} and related approaches~\cite{sun2023next3d} incorporate explicit 3D representations into the generative pipeline, but require 3D-aware training data or specialized architectures that limit generalization. Score Distillation Sampling (SDS)~\cite{poole2022dreamfusion} and its variants~\cite{tang2023dreamgaussian} offer a training-free alternative by using 2D diffusion priors to optimize 3D representations, though they are prone to over-saturation and mode collapse. Our method extends this to a bidirectional coupling: the diffusion model and 3DGS iteratively inform each other within the denoising loop, jointly improving photorealism and geometric consistency.

\begin{wrapfigure}{r}{0.6\textwidth}
    \vspace{-10pt}
    \centering
    \includegraphics[width=0.58\textwidth]{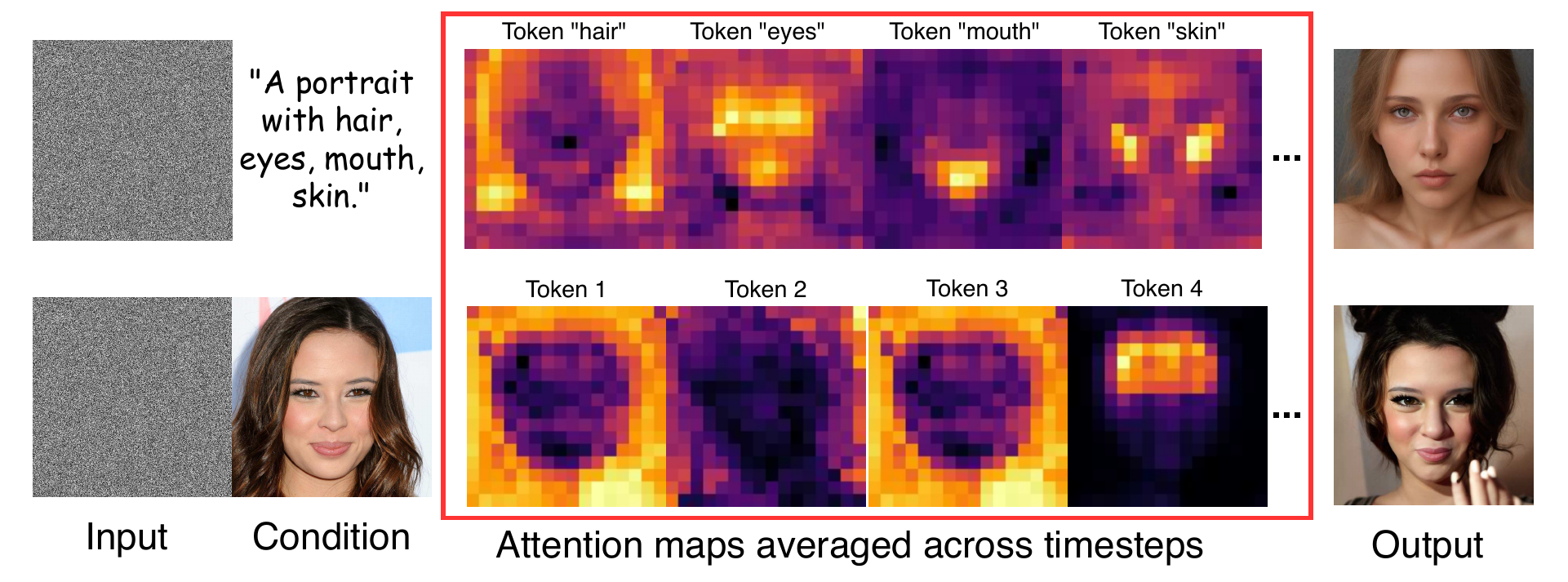}
    \caption{\textbf{Cross-attention maps.} Text tokens (top) activate over semantically distinct regions; image tokens (bottom) overlap broadly without spatial disentanglement.}
    \label{fig:attention}
    \vspace{-10pt}
\end{wrapfigure}

A parallel line of work edits existing 3DGS scenes by manipulating cross-attention maps to enforce multi-view consistency~\cite{wen2025intergsedit}. These methods inverse-render 2D attention maps onto 3D Gaussians and re-project them, ensuring different views attend to the same content at consistent locations. This is effective for text-guided editing, where tokens are semantically disentangled (e.g., ``hair'' activates over hair, ``eyes'' over eyes; Fig.~\ref{fig:attention}, top) and edits preserve the base geometry. However, identity-conditioned generation breaks both assumptions. Image-based conditioning tokens~\cite{ye2023ip} lack disentangled semantics and activate broadly across the entire face (Fig.~\ref{fig:attention}, bottom), causing shared attention to blend incompatible, view-dependent information. Moreover, transferring identity requires geometric deformation that fixed-geometry attention sharing cannot accommodate. This motivates our explicit 3D feedback loop operating in image space, allowing both appearance and geometry to evolve during denoising.

\section{Preliminaries}
\label{sec:prelim}
\textbf{3D Gaussian Splatting.} A 3D Gaussian Splatting (3DGS) model $\mathcal{M}$ represents a scene using $K$ anisotropic Gaussians $\{G_k\}$~\cite{kerbl20233d}. Each Gaussian has 59 parameters: center $\mu_k\in\mathbb{R}^3$, opacity $\alpha_k\in[0,1]$, 48 spherical harmonic (SH) color coefficients $c_k$, and a covariance matrix $\Sigma_k=R_k S_k^2 R_k^T$ (factored into a rotation $R_k$ and diagonal scale $S_k$). The framework performs differentiable rasterization to render an image from camera parameters (extrinsics and intrinsics) $v$: $\hat{I} = \mathcal{R}(\mathcal{M}, v)$, where $\mathcal{R}(\cdot, \cdot)$ alpha-blends the colors of Gaussians intersecting each cast ray: \mbox{$C=\sum_{d=1}^{D} c_d\,\alpha_d \prod_{j<d}(1-\alpha_j)$}. The model's parameters are optimized by minimizing a standard photometric loss between rendered and target images:
\begin{equation}
    \mathcal{L}_\mathrm{photo}(I, \hat{I}) = \lambda_1 \| I - \hat{I} \|_1 + \lambda_2 (1 - \mathrm{SSIM}(I, \hat{I})),
    \label{eq:photo}
\end{equation}
where $\lambda_1$ and $\lambda_2$ are weighting coefficients.

\textbf{Denoising Diffusion Implicit Models (DDIMs).} DDIMs~\cite{song2020denoising} define a deterministic sampling process over a learned reverse diffusion trajectory with $T$ total timesteps. Starting from a clean image $\mathbf{x}_0$, the forward process produces a noisy latent at any timestep $t \in \{1, \ldots, T\}$:
\begin{equation}
    \mathbf{x}_t = \alpha_t \mathbf{x}_0 + \sigma_t \boldsymbol{\epsilon}_\mathrm{gt}, \quad \boldsymbol{\epsilon}_\mathrm{gt} \sim \mathcal{N}(\mathbf{0}, \mathbf{I}),
    \label{eq:forward}
\end{equation}
where $\alpha_t$ and $\sigma_t$ are noise schedule parameters. In the reverse process, a noise prediction network 
$\boldsymbol{\epsilon}_\theta(\cdot, \cdot, \cdot)$ estimates the noise at each step: $\boldsymbol{\epsilon}_{\theta}^t = \boldsymbol{\epsilon}_\theta(\mathbf{x}_t, t, \mathbf{c})$ conditioned on context $\mathbf{c}$. We get the predicted clean image by:
\begin{equation}
    \hat{\mathbf{x}}_0 = \frac{\mathbf{x}_t - \sigma_t \boldsymbol{\epsilon}_{\theta}^t}{\alpha_t},
    \label{eq:ddim_x0}
\end{equation}
and the image at the previous timestep by:
\begin{equation}
    \mathbf{x}_{t-1} = \alpha_{t-1} \hat{\mathbf{x}}_0 + \sigma_{t-1} \, \boldsymbol{\epsilon}_{\theta}^t.
    \label{eq:ddim_step}
\end{equation}
For latent diffusion models \cite{rombach2022high}, this entire process is performed on latent codes $\mathbf{x}_t$ rather than on pixels, defined by an encoder $\mathcal{E}(\cdot)$ and decoder $\mathcal{D}(\cdot)$ that map images to and from the latent space.
\vspace{-0.2cm}
\section{Method}
\vspace{-0.2cm}
\label{sec:method}

\begin{figure*}[!t]
    \centering
   \includegraphics[width=\textwidth]{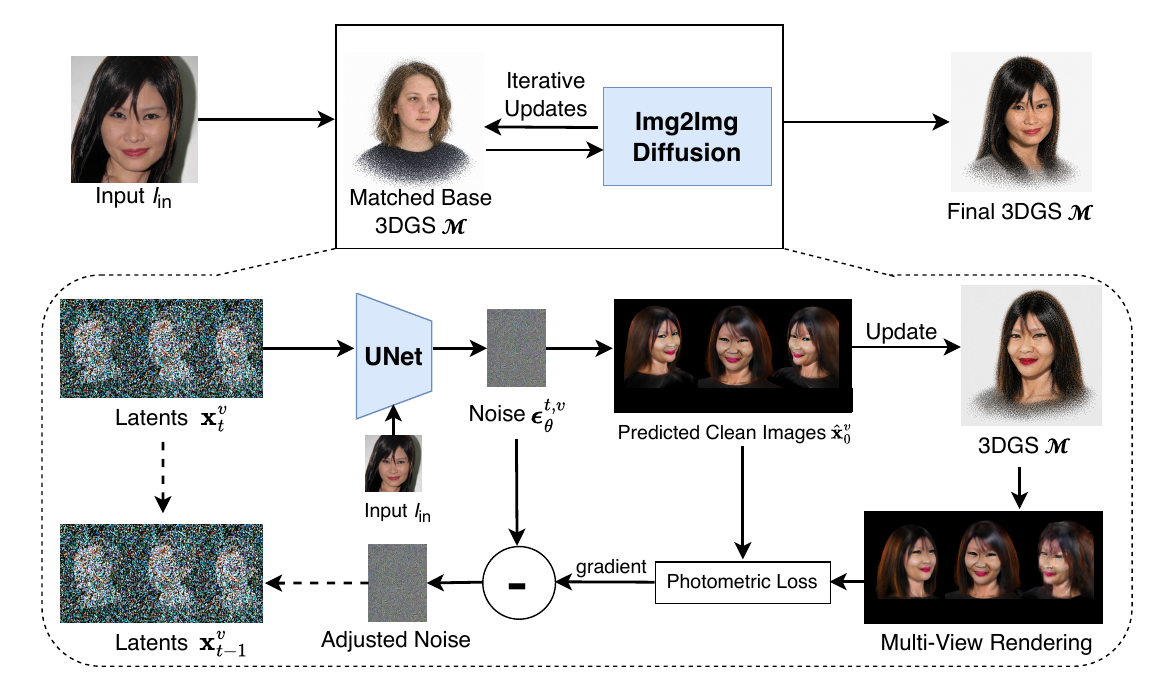}
   \vspace{-0.5cm}
   \caption{\textbf{Method overview.} \textbf{(Top)} Given an input photograph $I_\mathrm{in}$, SplatShot selects a matching base 3DGS model $\mathcal{M}$ and iteratively refines it through a 3DGS-guided img2img diffusion process, producing a final 3DGS avatar. \textbf{(Bottom)} At each denoising step, the UNet predicts per-view noise $\boldsymbol{\epsilon}_\theta^{t,v}$ conditioned on $I_\mathrm{in}$, from which predicted clean images $\hat{\mathbf{x}}_0^v$ are decoded. These images are used to update $\mathcal{M}$, which is then re-rendered from all views. The photometric loss $\mathbf{g}$ between the predicted images and the 3DGS re-renderings is backpropagated to adjust the noise, and the adjusted noise produces the latents for the next timestep.}
   \label{fig:method}
\end{figure*}

Given a single unconstrained input photograph $I_\mathrm{in}$, our goal is to produce a 3DGS model that captures the depicted identity and renders photorealistically from novel viewpoints. We assume access to a set of pretrained ``base'' 3DGS face models, a latent diffusion model comprising an encoder $\mathcal{E}$, decoder $\mathcal{D}$, and noise prediction network $\boldsymbol{\epsilon}_\theta$, as well as a differentiable 3DGS renderer $\mathcal{R}$ (Sec.~\ref{sec:prelim}).

\alg{} first automatically selects the base model $\mathcal{M}$ whose facial attributes (e.g., face shape, skin tone, hairstyle) best match $I_\mathrm{in}$ to provide a structured initialization to the optimization process (see Appendix for details on our selection scheme). Using $\mathcal{M}$, we render $V$ images from different views and proceed with an iterative, three-step scheme: (1) denoise each view using the diffusion model to match the input identity, (2) refit $\mathcal{M}$ to these denoised images and re-render them, and (3) back-propagate the photometric discrepancy between the 3D re-renderings and the 2D diffusion predictions into the noise space. Through this joint optimization, the diffusion model is steered toward 3D-consistent outputs while the 3DGS model converges on the target identity, mutually reinforcing photorealism and geometric strictness.

We show the method overview in Figure \ref{fig:method}, and detail this iterative framework in \S\ref{sec:reverse_sds}, and a crucial strategy to encourage stable early-step guidance in \S\ref{sec:hybrid}.
\vspace{-0.13cm}
\subsection{3DGS-Guided Iterative Denoising Framework}
\vspace{-0.13cm}
\label{sec:reverse_sds}
Let us denote the set of input views from base model $\mathcal{M}$ by $\{I^v\}$. We first encode these images into the diffusion model's latent space via $\mathcal{E}$ to yield a set of latent codes $\{\mathbf{x}_0^v\}$. We then add noise following Eq.~\ref{eq:forward} up to a starting timestep $t_s = \lfloor s \cdot T \rfloor$, where $s \in (0, 1]$. Starting from partial noise rather than pure noise ($t_s < T$) preserves coarse structural cues from the base model's renderings, providing a more stable initialization for the 3DGS guidance loop. 
The following denoising procedure proceeds from $t = t_s$ to $t = 0$.
 
\textbf{Step 1: Per-View Image Generation with Diffusion.} At timestep $t$, we independently predict noise for each view 
$\boldsymbol{\epsilon}_{\theta}^{t,v}=\boldsymbol{\epsilon}_\theta(\mathbf{x}_t^v, t, \mathbf{c})$, where $\mathbf{c}$ encodes the identity from $I_\mathrm{in}$ via an image-based adapter (e.g., IP-Adapter~\cite{ye2023ip}). 
We then recover the predicted clean latent $\hat{\mathbf{x}}_0^v$ via Eq.~\ref{eq:ddim_x0} and decode the corresponding image: $\hat{I}^v = \mathcal{D}(\hat{\mathbf{x}}_0^v)$. Because we process each view independently, these images lack explicit cross-view consistency.
 
\textbf{Step 2: 3DGS Model Update.} We update all Gaussian parameters of $\mathcal{M}$ with $\{\hat{I}^v\}$ for $N$ gradient steps using the loss function in Eq.~\ref{eq:photo}. Fine-tuning $\mathcal{M}$ across timesteps in this manner allows facial structure to evolve progressively and naturally resolve per-view inconsistencies into a geometrically plausible consensus rather than reproducing artifacts of any single view.

\textbf{Step 3: 3DGS-Guided Noise Adjustment.} Finally, we use the fine-tuned $\mathcal{M}$ to steer the predicted per-view noises towards image sets with 3D consistency. Let $\mathbf{g} = \mathcal{L}_\mathrm{photo}(\hat{I}^v,\, \mathcal{R}(\mathcal{M}, v))$ compute the loss between the diffusion-generated views from Step 1 and current $\mathcal{M}$-rendered views. We compute per-view noise adjustments via:
\begin{equation}
\boldsymbol{\epsilon}^{t,v}_\theta \leftarrow \boldsymbol{\epsilon}^{t,v}_\theta - \lambda \cdot  \frac{\partial \mathbf{g}}{\partial \boldsymbol{\epsilon}^{t,v}_\theta},
    \label{eq:noise_adjust}
\end{equation}
where $\lambda$ is a hyperparameter. The gradient $\partial \mathbf{g} / \partial \boldsymbol{\epsilon}^{t,v}_\theta$ can be computed via the chain rule:
\begin{equation}
\frac{\partial \mathbf{g}}{\partial \boldsymbol{\epsilon}^{t,v}_\theta} = \frac{\partial \hat{\mathbf{x}}_0^v}{\partial \boldsymbol{\epsilon}^{t,v}_\theta} \cdot \frac{\partial \mathbf{g}}{\partial \hat{\mathbf{x}}_0^v},
    \label{eq:chain}
\end{equation}
where $\partial \hat{\mathbf{x}}_0^v / \partial \boldsymbol{\epsilon}^{t,v}_\theta = -\sigma_t / \alpha_t$ from Eq.~\ref{eq:ddim_x0} and $\partial \mathbf{g} / \partial \hat{\mathbf{x}}_0^v$ may be computed via auto-differentiation through $\mathcal{D}$. Finally, we compute the next latents $\mathbf{x}_{t-1}^v$ using this adjusted noise via Eq.~\ref{eq:ddim_step} and continue the DDIM denoising chain moving back to Step 1. And at the end of the diffusion process (time step $t=0$), we can render the final views using $\mathcal{M}$.

\begin{figure}[t!]
    \centering
   \includegraphics[width=0.95\textwidth]{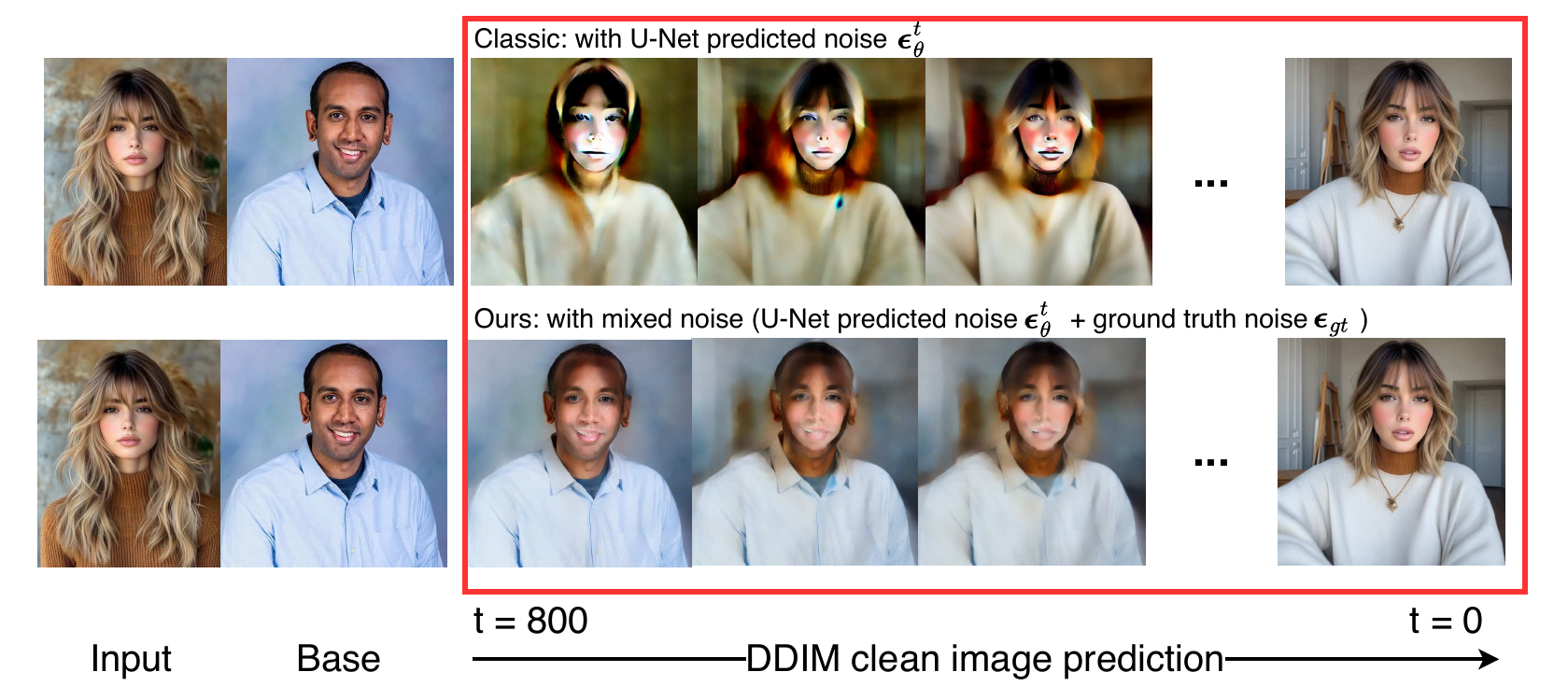}
   \vspace{-0.2cm}
   \caption{\textbf{Noise composition affects the structural progression of generated faces during the diffusion denoising process (see \S~\ref{sec:hybrid}).} Top: using predicted noise alone yields corrupted early-step predictions with over-saturated colors and unstable backgrounds. Bottom (Ours): our noise mixture mechanism (Eq.~\ref{eq:hybrid_noise}) blends predicted and ground-truth noise, producing stable, identity-consistent predictions from the first step, enabling reliable 3DGS fitting throughout the denoising process.}
   \vspace{-0.2cm}
   \label{fig:hybrid}
\end{figure}

\subsection{Noise Mixture for Early-Step Stability}
\label{sec:hybrid}
The above procedure assumes that the generated views $\hat{I}^v$ from Step 1 are good signals to supervise 3DGS fitting. However, we find that at early denoising steps (large $t$), the predicted $\hat{\mathbf{x}}_0$ from Eq.~\ref{eq:ddim_x0} is severely corrupted with over-saturated colors and incoherent backgrounds (see Fig.~\ref{fig:hybrid}, top). Refitting $\mathcal{M}$ to such targets causes catastrophic, unrecoverable shifts to Gaussian parameters, derailing convergence.

Because we control the forward noising process, we have access to the ground-truth noise $\boldsymbol{\epsilon}_\mathrm{gt}$ added during initialization (Eq.~\ref{eq:forward}). Substituting $\boldsymbol{\epsilon}_\mathrm{gt}$ into Eq.~\ref{eq:ddim_x0} recovers substantially cleaner images, but breaks the computational graph because $\boldsymbol{\epsilon}_\mathrm{gt}$ carries no gradient from $\boldsymbol{\epsilon}_\theta$. To maintain both stability and differentiability, we mix the two noises:
\begin{equation}
\boldsymbol{\epsilon}_\mathrm{mix}^{t,v} = w \cdot \boldsymbol{\epsilon}_\theta^{t,v} + (1 - w) \cdot \boldsymbol{\epsilon}_\mathrm{gt}^v,
    \label{eq:hybrid_noise}
\end{equation}
where $w \in [0, 1]$. We then obtain the hybrid clean prediction by substituting $\boldsymbol{\epsilon}_\mathrm{mix}^v$ into Eq.~\ref{eq:ddim_x0}. The predicted noise component preserves the gradient flow required for geometry guidance (Eq.~\ref{eq:chain}), while the ground-truth component ensures a smooth transition that prevents 3DGS collapse (Fig.~\ref{fig:hybrid}, bottom).

\begin{figure*}[!t]
    \centering
   \includegraphics[width=1.01\textwidth]{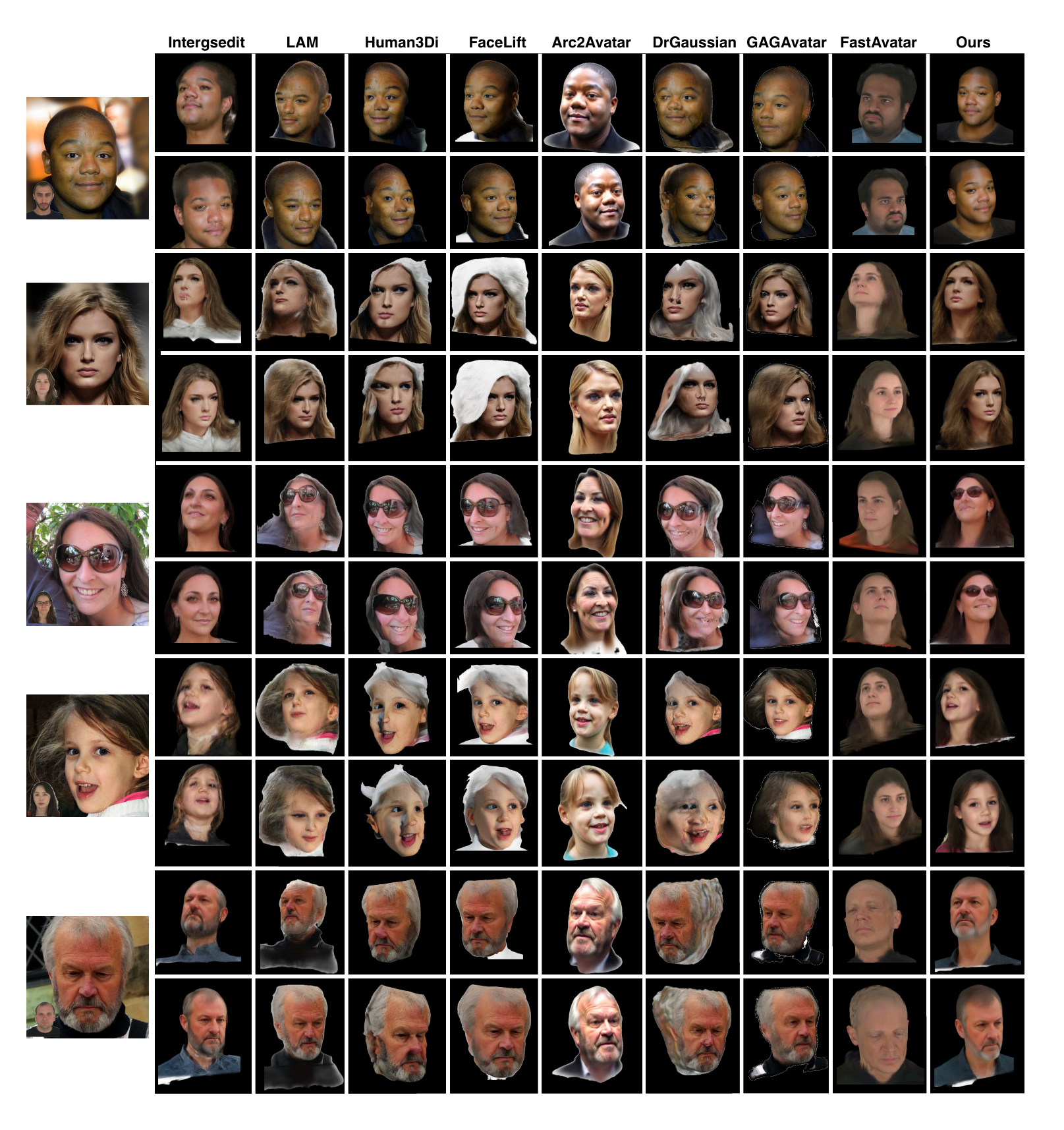}
   \vspace{-0.6cm}
   \caption{\textbf{Qualitative comparison on 3D face avatar generation from a single image.} Given a single unconstrained input (left), we compare our method with Intergsedit \cite{wen2025intergsedit}, LAM~\cite{he2025lam}, Human-3Diffusion~\cite{xue2024human}, FaceLift~\cite{lyu2025facelift}, Arc2Avatar~\cite{gerogiannis2025arc2avatar}, DreamGaussian~\cite{tang2023dreamgaussian}, GAGAvatar~\cite{chu2024generalizable}, and FastAvatar~\cite{liang2025fastavatar}. Previous methods yield synthetic-looking results, struggle with out-of-distribution inputs, produce low-quality novel views, or suffer from identity mismatch. In contrast, \alg{} renderings maintain identity and geometric consistency across views. The base 3DGS models for Intergsedit and Ours are shown in the bottom-left corner of each input. Additional examples are in Appendix.}
   \vspace{-0.5cm}
   \label{fig:main_results}
\end{figure*}

\vspace{-0.13cm}
\section{Experiments}
\label{sec:experiments}
\vspace{-0.13cm}
We evaluated \alg on single-view 3D face reconstruction using 6,279 randomly selected images from the CelebA~\cite{liu2015faceattributes} and FFHQ~\cite{karras2019style} datasets, capturing a diverse distribution of ages, ethnicities, poses, lighting conditions, and accessories. We collected 300 base 3DGS models ($\mathcal{M}$) pretrained on multi-view sequences from the NeRSemble dataset~\cite{kirschstein2023nersemble}.

\textbf{Baselines.} We compare against representative methods from three categories: 
(1) \emph{single-image to 3DGS}: 
LAM~\cite{he2025lam}, DreamGaussian~\cite{tang2023dreamgaussian}, GAGAvatar~\cite{chu2024generalizable}, FastAvatar~\cite{liang2025fastavatar}.
(2) \emph{diffusion-based methods}:
Human-3Diffusion~\cite{xue2024human}, FaceLift~\cite{lyu2025facelift}, Arc2Avatar~\cite{gerogiannis2025arc2avatar}.
(3) \emph{3DGS editing}: Intergsedit~\cite{wen2025intergsedit} (while Intergsedit primarily focuses on text-based editing, we re-implement on top of their framework for image-based editing. See \S\ref{sec:supp_gaussianedit} for details). 

\textbf{Implementation details.} We use Stable Diffusion v1.5 as the base diffusion model, augmented with IP-Adapter-Plus-Face~\cite{ye2023ip} for identity conditioning and ControlNet~\cite{zhang2023adding} for pose control. The 3DGS model is implemented using gsplat~\cite{ye2025gsplat}. We render $V = 16$ views from the base model. During geometry-guided denoising, we perform 50 DDIM steps with a strength of $s = 0.6$. The 3DGS refit uses 1,000 iterations at the first guidance step and 200 at subsequent steps. We set the noise mixture weight to $w = 0.4$ and the 3DGS-guidance weight to $\lambda = 50{,}000$. All experiments are conducted on a single NVIDIA A100 GPU. A detailed implementation is in Appendix \S\ref{sec:impl_details}.


\textbf{Metrics.} We consider several metrics across different dimensions. 
\textbf{Identity fidelity}: CSIM measures ArcFace~\cite{deng2018arcface} cosine similarity between the reference image and near-frontal generated views. AKD measures the average normalized 3D landmark distance estimated by DECA~\cite{feng2021deca}. 
\textbf{3D consistency}: CV-CSIM measures pairwise ArcFace~\cite{deng2018arcface} similarity across generated views. AED measures expression coefficient drift estimated by DECA~\cite{feng2021deca}. 
\textbf{Photo quality}: We report FID~\cite{heusel2017gans} on near-frontal views against FFHQ, and CLIP-IQA~\cite{wang2023exploring} as a no-reference perceptual quality score. 
(Because our outputs are rendered avatars rather than natural photographs, FID is interpreted comparatively rather than as an absolute photorealism measure.)

\subsection{Qualitative Comparison}
Fig.~\ref{fig:main_results} presents qualitative comparisons across five identities with varying appearances, poses, ages, and occlusions. We present two novel views per identity to assess visual quality and multi-view consistency. Intergsedit exhibits limited identity control under image-based guidance and produces blurry novel views. Diffusion-based baselines (Arc2Avatar, FaceLift, Human-3Diffusion) often generate plausible individual images but suffer from identity drift and inconsistent details across views. Feed-forward 3DGS methods (LAM, GAGAvatar, DreamGaussian, FastAvatar) produce explicit 3D outputs but exhibit visible artifacts, synthetic-looking textures, or identity mismatch on challenging in-the-wild inputs.

In contrast, \alg{} produces sharper, highly identity-faithful 3DGS renderings, even under challenging conditions like occlusions, extreme ages, diverse skin tones, and non-frontal poses. The novel views remain consistent in facial structure and appearance, demonstrating the advantage of optimizing an explicit 3DGS avatar as the final output. We show additional qualitative results and multi-view outputs from diffusion in the Appendix.

\subsection{Quantitative Comparison}
\label{sec:quant}

Table~\ref{tab:quantitative} reports quantitative results on 100 FFHQ identities (16 rendered views each). \alg{} achieves best or highly competitive performance across all metrics. It secures the highest CV-CSIM, lowest AKD, lowest FID, and highest CLIP-IQA, indicating superior cross-view consistency, accurate facial geometry, and excellent perceptual quality. It also ranks second in CSIM and remains competitive in AED, confirming strong reference identity preservation and expression stability. Compared to feed-forward 3DGS methods, \alg{} dramatically improves photo quality and multi-view consistency while maintaining comparable identity fidelity. Compared to diffusion baselines, \alg{} provides far stronger cross-view consistency.

\begin{table}[t]
\centering
\caption{\textbf{Quantitative comparison of various methods on FFHQ (100 identities, 16 views each).} We use colors to denote the \colorbox{gold}{\textbf{first}} and \colorbox{silver}{second} places. \alg{} achieves best or highly competitive performance across all metrics.}
\label{tab:quantitative}
\small
\setlength{\tabcolsep}{3pt}
\begin{tabular}{l cccccc}
\toprule
Method & CSIM$\uparrow$ & CV-CSIM$\uparrow$ & AKD$(\times 10^{-2})\downarrow$ & AED$\downarrow$ & FID$\downarrow$ & CLIP-IQA$\uparrow$ \\
\midrule
Intergsedit \cite{wen2025intergsedit} & .525 & .592 & 1.96 & .285 & 288 & .239 \\
LAM~\cite{he2025lam} & .584 & .623 & 1.80 & .270 & 277 & .369 \\
Human-3Di~\cite{xue2024human} & .481 & .546 & 1.20 & \first{.179} & 268 & .399 \\
FaceLift~\cite{lyu2025facelift} & .648 & .664 & 1.22 & .246 & 242 & .399 \\
Arc2Avatar~\cite{gerogiannis2025arc2avatar} & .608 & .801 & 1.12 & .240 & 257 & \second{.621} \\
DreamGaussian~\cite{tang2023dreamgaussian} & .663 & .661 & 2.16 & .276 & 262 & .290 \\
GAGAvatar~\cite{chu2024generalizable} & \first{.701} & .577 & \second{1.07} & \second{.190} & 246 & .566 \\
FastAvatar~\cite{liang2025fastavatar} & .345 & \second{.811} & 1.85 & .266 & \second{230} & .549 \\
\midrule
\alg{} (ours) & \second{.698} & \first{.832} & \first{0.92} & .202 & \first{216} & \first{.633} \\
\bottomrule
\end{tabular}
\end{table}
\subsection{Ablation Study}

\begin{figure}[t]
    \centering
    \includegraphics[width=\textwidth]{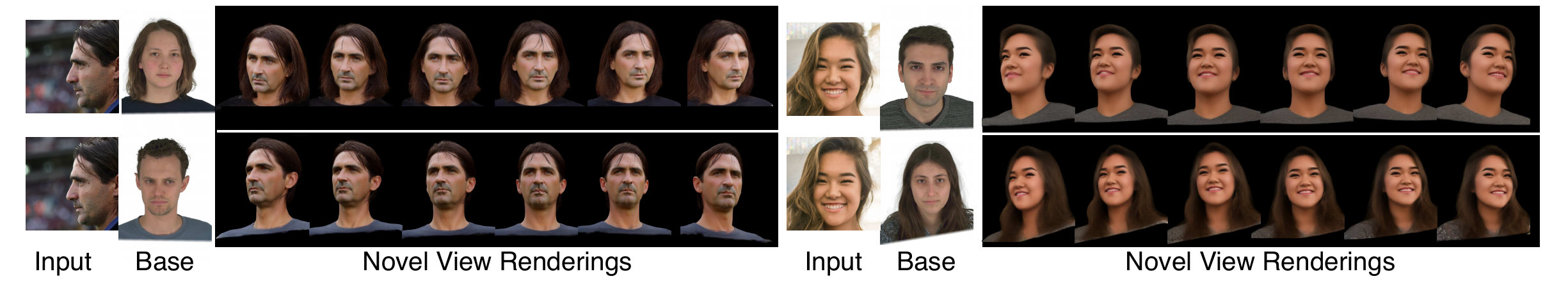}
    \vspace{-0.5cm}
    \caption{\textbf{Effect of base 3DGS model selection.} Each pair of rows shows the same input generated with two different base models. Identity is successfully transferred in both cases, but hairstyle and head shape are largely inherited from the base. This is expected: trained identity encoders typically crop out hair, and hair geometry is inherently less stable for 3DGS reconstruction. We therefore prioritize hairstyle matching in our base selection algorithm (\S\ref{sec:hair_alg}). More results in Fig.\ref{fig:same_base}, \ref{fig:same_ref}}
    \label{fig:template}
    \vspace{-0.2cm}
\end{figure}

\begin{figure}[h]
    \centering
    \includegraphics[width=\textwidth]{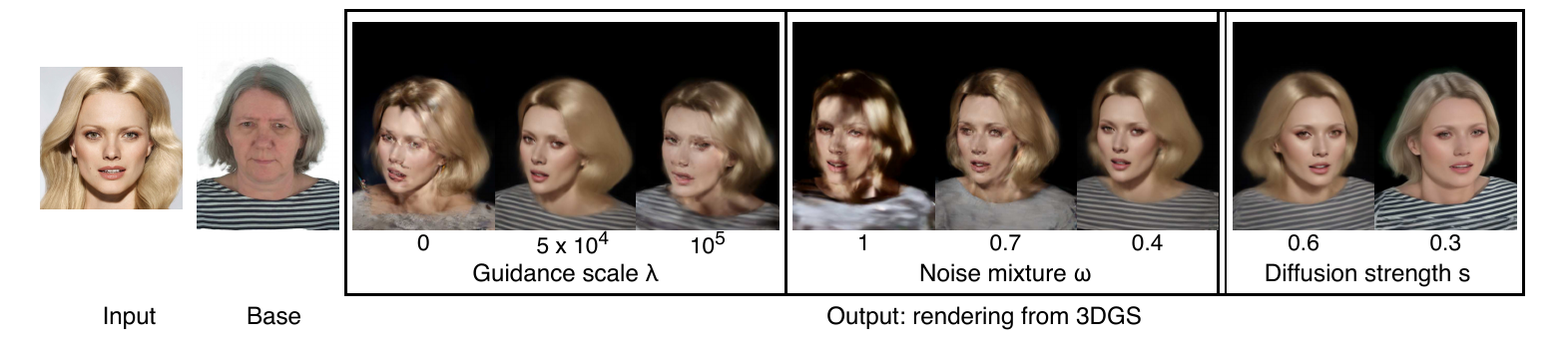}
    \vspace{-0.5cm}
    \caption{\textbf{Visual examples of ablation study.} Left to right: effect of guidance scale $\lambda$, hybrid weight $w$, and denoising strength $s$. Without guidance ($\lambda = 0$), 3DGS rendering lacks 3D consistency in novel views. Excessive guidance ($\lambda = 100{,}000$) over-constrains the output. Without hybrid prediction ($w = 1$), early inconsistent predictions force Gaussians to compensate with unstable positions and colors, degrading rendering quality. Lower $w$ stabilizes the predictions. Lower strength ($s = 0.3$) preserves the base geometry and appearance with limited identity transfer, while higher strength ($s = 0.6$) allows stronger transfer including changes to facial structure.}
    \label{fig:ablation}
    
\end{figure}

Table~\ref{tab:ablation} ablates key design choices across 20 identities (16 views each), evaluating identity fidelity (CSIM), 3D consistency (CV-CSIM), and perceptual quality (CLIP-IQA). We provide visual examples in Fig.~\ref{fig:ablation}. Without geometry guidance ($\lambda = 0$), CSIM remains moderate but CV-CSIM drops sharply, showing that the 3DGS feedback loop is vital for cross-view coherence. Setting $\lambda = 5 \times 10^4$ provides the optimal balance; excessive guidance over-constrains the diffusion output and degrades all metrics. Relying solely on predicted noise ($w = 1$) forces Gaussians to compensate for inconsistent early-step predictions, yielding unstable geometry (Fig.~\ref{fig:ablation}, middle). Lowering $w$ stabilizes these predictions by blending in ground-truth noise, with $w = 0.4$ performing best. The denoising strength ($s$) dictates the balance between identity transfer and base preservation: lower values retain the base appearance, while higher values (e.g., $s=0.6$) enable strong identity transfer at a slight consistency trade-off. Using $V=16$ views offers a strong trade-off for consistency, and increasing refit steps $M$ beyond 200 yields diminishing returns. Removing ControlNet pose conditioning degrades both identity and consistency, as the generated faces drift from their intended viewpoints.

\begin{table}[h]
\centering
\caption{\textbf{Ablation study on key design choices.} * denotes default. \colorbox{gold}{\textbf{Best}} per group is highlighted.}
\label{tab:ablation}
\small
\setlength{\tabcolsep}{2pt}
\begin{tabular}{l ccccc ccccc cccc ccc cc}
\toprule
Term & \multicolumn{4}{c}{Guidance $\lambda$} & \multicolumn{4}{c}{Hybrid $w$} & \multicolumn{3}{c}{Strength $s$} & \multicolumn{3}{c}{Views $V$} & \multicolumn{3}{c}{Refit $M$} & \multicolumn{2}{c}{CtrlNet} \\
\cmidrule(lr){1-1} \cmidrule(lr){2-5} \cmidrule(lr){6-9} \cmidrule(lr){10-12} \cmidrule(lr){13-15} \cmidrule(lr){16-18} \cmidrule(lr){19-20}
Setting & $0$ & $10^4$ & $5\!\times\!10^4$* & $10^5$ & $1$ & $.7$ & $.4$* & $.1$ & $.3$ & $.6$* & $.8$ & $8$ & $16$* & $32$ & $50$ & $200$* & $500$ & w/o & w/* \\
\midrule
CSIM$\uparrow$ & .56 & .60 & \first{.70} & .64 & .64 & .69 & \first{.70} & .69 & .41 & .70 & \first{.74} & .68 & \first{.70} & .67 & .64 & .70 & \first{.73} & .65 & \first{.70} \\
CV-CSIM$\uparrow$ & .38 & .57 & .83 & \first{.84} & .82 & .82 & \first{.83} & .81 & \first{.89} & .83 & .74 & .61 & .83 & \first{.86} & .69 & \first{.83} & \first{.83} & .74 & \first{.83} \\
CLIP-IQA$\uparrow$ & .45 & .52 & \first{.63} & .61 & .39 & .48 & .63 & \first{.63} & \first{.67} & .63 & .56 & .59 & \first{.63} & .60 & .52 & .63 & \first{.65} & .45 & \first{.63} \\
\bottomrule
\end{tabular}
\end{table}

We additionally show the effect of varying the base model in Fig.~\ref{fig:template}: the same input with different bases produces consistent identity but with hairstyle largely inherited from each base, as trained identity encoders typically crop out hair and hair geometry is inherently challenging for 3DGS. This motivates our base selection algorithm, which prioritizes hairstyle matching (details in \S\ref{sec:hair_alg}). The denoising strength $s$ controls also how much the output departs from the base geometry: lower $s$ preserves the base appearance while higher $s$ allows the facial structure itself to change (Fig.~\ref{fig:ablation}), confirming that the base model serves as a soft anchor rather than a rigid constraint. We show more results and a detailed analysis in \S\ref{sec:supp_base}, Fig. \ref{fig:same_base}, \ref{fig:same_ref}.

\vspace{-0.2cm}
\section{Discussion and Conclusion}
\label{sec:conclusion}

Our results suggest that unconstrained single-image avatar generation benefits from combining image-space generative priors with an explicit 3D representation. Feed-forward 3D avatar methods such as GAGAvatar, FastAvatar, and LAM can produce stable outputs when the input is close to their training distribution, but often struggle with in-the-wild photos that differ in pose, lighting, accessories, or capture quality. Diffusion-based methods such as FaceLift, Arc2Avatar, and Human-3Diffusion benefit from strong 2D image priors and can synthesize plausible individual views, but these views are not always explained by a single coherent 3D representation. In contrast, \alg{} uses diffusion to improve realism while repeatedly projecting multi-view predictions back into an explicit 3DGS avatar, leading to renderings that better preserve identity and remain consistent across novel views. Additional challenging cases, including incomplete captures and large pose or lighting variations, are provided in the Appendix. Notably, although GAGAvatar obtains strong CSIM in our evaluation, its high-fidelity outputs rely on a neural rendering stage; when these outputs are refit into a 3DGS, severe artifacts appear, as shown in Fig.~\ref{fig:novel_view_refit}.

\begin{wraptable}{r}{0.3\textwidth}
\vspace{-12pt}
\centering
\label{tab:demographic}
\small
\begin{tabular}{l c}
\toprule
Demographic & CSIM$\uparrow$ \\
\midrule
Caucasian & .72 \\
African & .69 \\
Asian & .69 \\
Hispanic & .70 \\
Children & .66 \\
Elderly & .67 \\
\midrule
Dataset & .70 \\
\bottomrule
\end{tabular}
\vspace{-12pt}
\end{wraptable}

\paragraph{Demographic Fairness.}
We further evaluate identity preservation across demographic subgroups in Table (right). \alg{} obtains similar CSIM scores across race and age groups, with only modest variation across subgroups. This suggests that the method does not strongly favor a single demographic group in our evaluation. We attribute this in part to using pretrained image generative priors and test-time 3DGS fitting rather than learning a feed-forward mapping solely from a curated 3D avatar dataset.

\paragraph{Limitations.}
Our method has several limitations. First, iterative 3DGS refitting makes the pipeline slower than feed-forward methods, taking approximately 3 minutes per identity on an A100 compared with seconds for GAGAvatar or FastAvatar. We explored flow-matching-based\cite{yan2024perflow} generation as a faster alternative, but found that it led to less consistent multi-view outputs; details and negative results are provided in the Appendix. Second, our geometry is anchored by base models trained on NeRSemble~\cite{kirschstein2023nersemble}, whose captures primarily cover the frontal 180-degree range; as a result, we have not yet evaluated our framework on back-view avatar generation. Additionally, since hairstyle is largely inherited from the base model rather than the input image, a poor hairstyle match between the input and the selected base can result in discrepancies. Finally, while our method handles common accessories such as sunglasses and earrings (Fig.~\ref{fig:main_results}), hats remain challenging. This is partly because NeRSemble lacks hat examples, and partly because the face identity encoder can entangle headwear with identity cues, treating hats as part of the person rather than a removable accessory. We show a failure case in Appendix Fig.~\ref{fig:celeba3d_results_01}.

\paragraph{Broader Impacts.}
By combining our method with off-the-shelf face detection~\cite{deng2020retinaface}, we build an end-to-end system that takes an unconstrained photograph, including a casual group photo (Fig.~\ref{fig:teaser}), and produces individual 3D avatars without additional input. This has applications in AR/VR, gaming, telepresence, and digital content creation, where scalable avatar generation from everyday photos could lower the barrier to personalized 3D experiences. However, the same capability may also enable unauthorized digital replicas, deepfakes, or identity misuse. We therefore recommend that deployments incorporate consent mechanisms. In our system implementation, we embed watermarks into images produced by our pipeline to support tracing and provenance verification.


{\small

\bibliographystyle{unsrtnat}

\bibliography{ref}

@article{kerbl20233d,
  title={3d gaussian splatting for real-time radiance field rendering.},
  author={Kerbl, Bernhard and Kopanas, Georgios and Leimk{\"u}hler, Thomas and Drettakis, George and others},
  journal={ACM Trans. Graph.},
  volume={42},
  number={4},
  pages={139--1},
  year={2023}
}

@article{song2020denoising,
  title={Denoising diffusion implicit models},
  author={Song, Jiaming and Meng, Chenlin and Ermon, Stefano},
  journal={arXiv preprint arXiv:2010.02502},
  year={2020}
}

@inproceedings{chen2024gaussianeditor,
  title={Gaussianeditor: Swift and controllable 3d editing with gaussian splatting},
  author={Chen, Yiwen and Chen, Zilong and Zhang, Chi and Wang, Feng and Yang, Xiaofeng and Wang, Yikai and Cai, Zhongang and Yang, Lei and Liu, Huaping and Lin, Guosheng},
  booktitle={Proceedings of the IEEE/CVF conference on computer vision and pattern recognition},
  pages={21476--21485},
  year={2024}
}

@inproceedings{wang2024gaussianeditor,
  title={Gaussianeditor: Editing 3d gaussians delicately with text instructions},
  author={Wang, Junjie and Fang, Jiemin and Zhang, Xiaopeng and Xie, Lingxi and Tian, Qi},
  booktitle={Proceedings of the IEEE/CVF conference on computer vision and pattern recognition},
  pages={20902--20911},
  year={2024}
}

@inproceedings{wang2024view,
  title={View-consistent 3d editing with gaussian splatting},
  author={Wang, Yuxuan and Yi, Xuanyu and Wu, Zike and Zhao, Na and Chen, Long and Zhang, Hanwang},
  booktitle={European conference on computer vision},
  pages={404--420},
  year={2024},
  organization={Springer}
}

@inproceedings{wen2025intergsedit,
  title={Intergsedit: Interactive 3d gaussian splatting editing with 3d geometry-consistent attention prior},
  author={Wen, Minghao and Wu, Shengjie and Wang, Kangkan and Liang, Dong},
  booktitle={Proceedings of the IEEE/CVF International Conference on Computer Vision},
  pages={26136--26145},
  year={2025}
}

@inproceedings{wu2024gaussctrl,
  title={Gaussctrl: Multi-view consistent text-driven 3d gaussian splatting editing},
  author={Wu, Jing and Bian, Jia-Wang and Li, Xinghui and Wang, Guangrun and Reid, Ian and Torr, Philip and Prisacariu, Victor Adrian},
  booktitle={European conference on computer vision},
  pages={55--71},
  year={2024},
  organization={Springer}
}

@inproceedings{chen2024dge,
  title={Dge: Direct gaussian 3d editing by consistent multi-view editing},
  author={Chen, Minghao and Laina, Iro and Vedaldi, Andrea},
  booktitle={European conference on computer vision},
  pages={74--92},
  year={2024},
  organization={Springer}
}

@article{zhuang2024tip,
  title={Tip-editor: An accurate 3d editor following both text-prompts and image-prompts},
  author={Zhuang, Jingyu and Kang, Di and Cao, Yan-Pei and Li, Guanbin and Lin, Liang and Shan, Ying},
  journal={ACM Transactions on Graphics (ToG)},
  volume={43},
  number={4},
  pages={1--12},
  year={2024},
  publisher={ACM New York, NY, USA}
}

@inproceedings{lee2025editsplat,
  title={Editsplat: Multi-view fusion and attention-guided optimization for view-consistent 3d scene editing with 3d gaussian splatting},
  author={Lee, Dong In and Park, Hyeongcheol and Seo, Jiyoung and Park, Eunbyung and Park, Hyunje and Baek, Ha Dam and Shin, Sangheon and Kim, Sangmin and Kim, Sangpil},
  booktitle={Proceedings of the Computer Vision and Pattern Recognition Conference},
  pages={11135--11145},
  year={2025}
}

@article{ye2023ip,
  title={Ip-adapter: Text compatible image prompt adapter for text-to-image diffusion models},
  author={Ye, Hu and Zhang, Jun and Liu, Sibo and Han, Xiao and Yang, Wei},
  journal={arXiv preprint arXiv:2308.06721},
  year={2023}
}

@inproceedings{wei2023elite,
  title={Elite: Encoding visual concepts into textual embeddings for customized text-to-image generation},
  author={Wei, Yuxiang and Zhang, Yabo and Ji, Zhilong and Bai, Jinfeng and Zhang, Lei and Zuo, Wangmeng},
  booktitle={Proceedings of the IEEE/CVF international conference on computer vision},
  pages={15943--15953},
  year={2023}
}

@article{wang2024instantid,
  title={Instantid: Zero-shot identity-preserving generation in seconds},
  author={Wang, Qixun and Bai, Xu and Wang, Haofan and Qin, Zekui and Chen, Anthony and Li, Huaxia and Tang, Xu and Hu, Yao},
  journal={arXiv preprint arXiv:2401.07519},
  year={2024}
}

@inproceedings{he2025lam,
  title={LAM: large avatar model for one-shot animatable gaussian head},
  author={He, Yisheng and Gu, Xiaodong and Ye, Xiaodan and Xu, Chao and Zhao, Zhengyi and Dong, Yuan and Yuan, Weihao and Dong, Zilong and Bo, Liefeng},
  booktitle={Proceedings of the Special Interest Group on Computer Graphics and Interactive Techniques Conference Conference Papers},
  pages={1--13},
  year={2025}
}

@article{xue2024human,
  title={Human-3Diffusion: realistic avatar creation via explicit 3D consistent diffusion models},
  author={Xue, Yuxuan and Xie, Xianghui and Marin, Riccardo and Pons-Moll, Gerard},
  journal={Advances in Neural Information Processing Systems},
  volume={37},
  pages={99601--99645},
  year={2024}
}

@inproceedings{lyu2025facelift,
  title={FaceLift: Learning generalizable single image 3D face reconstruction from synthetic heads},
  author={Lyu, Weijie and Zhou, Yi and Yang, Ming-Hsuan and Shu, Zhixin},
  booktitle={Proceedings of the IEEE/CVF International Conference on Computer Vision},
  pages={12691--12701},
  year={2025}
}

@inproceedings{gerogiannis2025arc2avatar,
  title={Arc2avatar: Generating expressive 3d avatars from a single image via id guidance},
  author={Gerogiannis, Dimitrios and Papantoniou, Foivos Paraperas and Potamias, Rolandos Alexandros and Lattas, Alexandros and Zafeiriou, Stefanos},
  booktitle={Proceedings of the Computer Vision and Pattern Recognition Conference},
  pages={10770--10782},
  year={2025}
}

@article{hertz2022prompt,
  title={Prompt-to-prompt image editing with cross attention control},
  author={Hertz, Amir and Mokady, Ron and Tenenbaum, Jay and Aberman, Kfir and Pritch, Yael and Cohen-Or, Daniel},
  journal={arXiv preprint arXiv:2208.01626},
  year={2022}
}

@article{poole2022dreamfusion,
  title={Dreamfusion: Text-to-3d using 2d diffusion},
  author={Poole, Ben and Jain, Ajay and Barron, Jonathan T and Mildenhall, Ben},
  journal={arXiv preprint arXiv:2209.14988},
  year={2022}
}

@article{liu2022flow,
  title={Flow straight and fast: Learning to generate and transfer data with rectified flow},
  author={Liu, Xingchao and Gong, Chengyue and Liu, Qiang},
  journal={arXiv preprint arXiv:2209.03003},
  year={2022}
}

@article{tang2023dreamgaussian,
  title={Dreamgaussian: Generative gaussian splatting for efficient 3d content creation},
  author={Tang, Jiaxiang and Ren, Jiawei and Zhou, Hang and Liu, Ziwei and Zeng, Gang},
  journal={arXiv preprint arXiv:2309.16653},
  year={2023}
}

@article{chu2024generalizable,
  title={Generalizable and animatable gaussian head avatar},
  author={Chu, Xuangeng and Harada, Tatsuya},
  journal={Advances in Neural Information Processing Systems},
  volume={37},
  pages={57642--57670},
  year={2024}
}

@article{liang2025fastavatar,
  title={FastAvatar: Instant 3D Gaussian Splatting for Faces from Single Unconstrained Poses},
  author={Liang, Hao and Ge, Zhixuan and Majee, Soumendu and Tiwari, Ashish and Godaliyadda, GM and Veeraraghavan, Ashok and Balakrishnan, Guha},
  journal={arXiv preprint arXiv:2508.18389},
  year={2025}
}

@article{kirschstein2023nersemble,
  title={Nersemble: Multi-view radiance field reconstruction of human heads},
  author={Kirschstein, Tobias and Qian, Shenhan and Giebenhain, Simon and Walter, Tim and Nie{\ss}ner, Matthias},
  journal={ACM Transactions on Graphics (TOG)},
  volume={42},
  number={4},
  pages={1--14},
  year={2023},
  publisher={ACM New York, NY, USA}
}

@inproceedings{liu2015faceattributes,
  title={Deep learning face attributes in the wild},
  author={Liu, Ziwei and Luo, Ping and Wang, Xiaogang and Tang, Xiaoou},
  booktitle={Proceedings of the IEEE international conference on computer vision},
  pages={3730--3738},
  year={2015}
}

@inproceedings{karras2019style,
  title={A style-based generator architecture for generative adversarial networks},
  author={Karras, Tero and Laine, Samuli and Aila, Timo},
  booktitle={Proceedings of the IEEE/CVF conference on computer vision and pattern recognition},
  pages={4401--4410},
  year={2019}
}

@inproceedings{zhang2023adding,
  title={Adding conditional control to text-to-image diffusion models},
  author={Zhang, Lvmin and Rao, Anyi and Agrawala, Maneesh},
  booktitle={Proceedings of the IEEE/CVF international conference on computer vision},
  pages={3836--3847},
  year={2023}
}

@article{ye2025gsplat,
  title={gsplat: An open-source library for Gaussian splatting},
  author={Ye, Vickie and Li, Ruilong and Kerr, Justin and Turkulainen, Matias and Yi, Brent and Pan, Zhuoyang and Seiskari, Otto and Ye, Jianbo and Hu, Jeffrey and Tancik, Matthew and Angjoo Kanazawa},
  journal={Journal of Machine Learning Research},
  volume={26},
  number={34},
  pages={1--17},
  year={2025}
}

@article{blanz2003face,
  title={Face recognition based on fitting a 3D morphable model},
  author={Blanz, Volker and Vetter, Thomas},
  journal={IEEE Transactions on pattern analysis and machine intelligence},
  volume={25},
  number={9},
  pages={1063--1074},
  year={2003},
  publisher={IEEE}
}

@inproceedings{smith2020morphable,
  title={A morphable face albedo model},
  author={Smith, William AP and Seck, Alassane and Dee, Hannah and Tiddeman, Bernard and Tenenbaum, Joshua B and Egger, Bernhard},
  booktitle={Proceedings of the IEEE/CVF Conference on Computer Vision and Pattern Recognition},
  pages={5011--5020},
  year={2020}
}

@incollection{blanz2023morphable,
  title={A morphable model for the synthesis of 3D faces},
  author={Blanz, Volker and Vetter, Thomas},
  booktitle={Seminal Graphics Papers: Pushing the Boundaries, Volume 2},
  pages={157--164},
  year={2023}
}

@inproceedings{li2020learning,
  title={Learning formation of physically-based face attributes},
  author={Li, Ruilong and Bladin, Karl and Zhao, Yajie and Chinara, Chinmay and Ingraham, Owen and Xiang, Pengda and Ren, Xinglei and Prasad, Pratusha and Kishore, Bipin and Xing, Jun and others},
  booktitle={Proceedings of the IEEE/CVF conference on computer vision and pattern recognition},
  pages={3410--3419},
  year={2020}
}

@article{mildenhall2021nerf,
  title={Nerf: Representing scenes as neural radiance fields for view synthesis},
  author={Mildenhall, Ben and Srinivasan, Pratul P and Tancik, Matthew and Barron, Jonathan T and Ramamoorthi, Ravi and Ng, Ren},
  journal={Communications of the ACM},
  volume={65},
  number={1},
  pages={99--106},
  year={2021},
  publisher={ACM New York, NY, USA}
}

@article{yan2024perflow,
  title={Perflow: Piecewise rectified flow as universal plug-and-play accelerator},
  author={Yan, Hanshu and Liu, Xingchao and Pan, Jiachun and Liew, Jun Hao and Liu, Qiang and Feng, Jiashi},
  journal={Advances in Neural Information Processing Systems},
  volume={37},
  pages={78630--78652},
  year={2024}
}

@inproceedings{rombach2022high,
  title={High-resolution image synthesis with latent diffusion models},
  author={Rombach, Robin and Blattmann, Andreas and Lorenz, Dominik and Esser, Patrick and Ommer, Bj{\"o}rn},
  booktitle={Proceedings of the IEEE/CVF conference on computer vision and pattern recognition},
  pages={10684--10695},
  year={2022}
}

@inproceedings{caron2021emerging,
  title={Emerging properties in self-supervised vision transformers},
  author={Caron, Mathilde and Touvron, Hugo and Misra, Ishan and J{\'e}gou, Herv{\'e} and Mairal, Julien and Bojanowski, Piotr and Joulin, Armand},
  booktitle={Proceedings of the IEEE/CVF international conference on computer vision},
  pages={9650--9660},
  year={2021}
}

@inproceedings{qian2024gaussianavatars,
  title={Gaussianavatars: Photorealistic head avatars with rigged 3d gaussians},
  author={Qian, Shenhan and Kirschstein, Tobias and Schoneveld, Liam and Davoli, Davide and Giebenhain, Simon and Nie{\ss}ner, Matthias},
  booktitle={Proceedings of the IEEE/CVF Conference on Computer Vision and Pattern Recognition},
  pages={20299--20309},
  year={2024}
}

@inproceedings{xiang2024flashavatar,
  title={Flashavatar: High-fidelity head avatar with efficient gaussian embedding},
  author={Xiang, Jun and Gao, Xuan and Guo, Yudong and Zhang, Juyong},
  booktitle={Proceedings of the IEEE/CVF Conference on Computer Vision and Pattern Recognition},
  pages={1802--1812},
  year={2024}
}

@inproceedings{deng2018arcface,
title={ArcFace: Additive Angular Margin Loss for Deep Face Recognition},
author={Deng, Jiankang and Guo, Jia and Niannan, Xue and Zafeiriou, Stefanos},
booktitle={CVPR},
year={2019}
}

@inproceedings{park2021nerfies,
  title={Nerfies: Deformable neural radiance fields},
  author={Park, Keunhong and Sinha, Utkarsh and Barron, Jonathan T and Bouaziz, Sofien and Goldman, Dan B and Seitz, Steven M and Martin-Brualla, Ricardo},
  booktitle={Proceedings of the IEEE/CVF international conference on computer vision},
  pages={5865--5874},
  year={2021}
}

@inproceedings{tretschk2021non,
  title={Non-rigid neural radiance fields: Reconstruction and novel view synthesis of a dynamic scene from monocular video},
  author={Tretschk, Edgar and Tewari, Ayush and Golyanik, Vladislav and Zollh{\"o}fer, Michael and Lassner, Christoph and Theobalt, Christian},
  booktitle={Proceedings of the IEEE/CVF international conference on computer vision},
  pages={12959--12970},
  year={2021}
}

@inproceedings{saunders2025gasp,
  title={GASP: Gaussian Avatars with Synthetic Priors},
  author={Saunders, Jack and Hewitt, Charlie and Jian, Yanan and Kowalski, Marek and Baltrusaitis, Tadas and Chen, Yiye and Cosker, Darren and Estellers, Virginia and Gyd{\'e}, Nicholas and Namboodiri, Vinay P and others},
  booktitle={Proceedings of the Computer Vision and Pattern Recognition Conference},
  pages={271--280},
  year={2025}
}

@inproceedings{chan2022efficient,
  title={Efficient geometry-aware 3d generative adversarial networks},
  author={Chan, Eric R and Lin, Connor Z and Chan, Matthew A and Nagano, Koki and Pan, Boxiao and De Mello, Shalini and Gallo, Orazio and Guibas, Leonidas J and Tremblay, Jonathan and Khamis, Sameh and others},
  booktitle={Proceedings of the IEEE/CVF conference on computer vision and pattern recognition},
  pages={16123--16133},
  year={2022}
}

@inproceedings{sun2023next3d,
  title={Next3d: Generative neural texture rasterization for 3d-aware head avatars},
  author={Sun, Jingxiang and Wang, Xuan and Wang, Lizhen and Li, Xiaoyu and Zhang, Yong and Zhang, Hongwen and Liu, Yebin},
  booktitle={Proceedings of the IEEE/CVF conference on computer vision and pattern recognition},
  pages={20991--21002},
  year={2023}
}

@inproceedings{zhuang2022mofanerf,
  title={Mofanerf: Morphable facial neural radiance field},
  author={Zhuang, Yiyu and Zhu, Hao and Sun, Xusen and Cao, Xun},
  booktitle={European conference on computer vision},
  pages={268--285},
  year={2022},
  organization={Springer}
}

@inproceedings{hong2022headnerf,
  title={Headnerf: A real-time nerf-based parametric head model},
  author={Hong, Yang and Peng, Bo and Xiao, Haiyao and Liu, Ligang and Zhang, Juyong},
  booktitle={Proceedings of the IEEE/CVF Conference on Computer Vision and Pattern Recognition},
  pages={20374--20384},
  year={2022}
}

@inproceedings{gafni2021dynamic,
  title={Dynamic neural radiance fields for monocular 4d facial avatar reconstruction},
  author={Gafni, Guy and Thies, Justus and Zollhofer, Michael and Nie{\ss}ner, Matthias},
  booktitle={Proceedings of the IEEE/CVF Conference on Computer Vision and Pattern Recognition},
  pages={8649--8658},
  year={2021}
}

@inproceedings{karras2020analyzing,
  title={Analyzing and improving the image quality of stylegan},
  author={Karras, Tero and Laine, Samuli and Aittala, Miika and Hellsten, Janne and Lehtinen, Jaakko and Aila, Timo},
  booktitle={Proceedings of the IEEE/CVF conference on computer vision and pattern recognition},
  pages={8110--8119},
  year={2020}
}

@inproceedings{an2023panohead,
  title={Panohead: Geometry-aware 3d full-head synthesis in 360deg},
  author={An, Sizhe and Xu, Hongyi and Shi, Yichun and Song, Guoxian and Ogras, Umit Y and Luo, Linjie},
  booktitle={Proceedings of the IEEE/CVF conference on computer vision and pattern recognition},
  pages={20950--20959},
  year={2023}
}

@inproceedings{shao2024splattingavatar,
  title={Splattingavatar: Realistic real-time human avatars with mesh-embedded gaussian splatting},
  author={Shao, Zhijing and Wang, Zhaolong and Li, Zhuang and Wang, Duotun and Lin, Xiangru and Zhang, Yu and Fan, Mingming and Wang, Zeyu},
  booktitle={Proceedings of the IEEE/CVF Conference on Computer Vision and Pattern Recognition},
  pages={1606--1616},
  year={2024}
}

@inproceedings{kocabas2024hugs,
  title={Hugs: Human gaussian splats},
  author={Kocabas, Muhammed and Chang, Jen-Hao Rick and Gabriel, James and Tuzel, Oncel and Ranjan, Anurag},
  booktitle={Proceedings of the IEEE/CVF conference on computer vision and pattern recognition},
  pages={505--515},
  year={2024}
}

@inproceedings{xu2024gaussian,
  title={Gaussian head avatar: Ultra high-fidelity head avatar via dynamic gaussians},
  author={Xu, Yuelang and Chen, Benwang and Li, Zhe and Zhang, Hongwen and Wang, Lizhen and Zheng, Zerong and Liu, Yebin},
  booktitle={Proceedings of the IEEE/CVF conference on computer vision and pattern recognition},
  pages={1931--1941},
  year={2024}
}

@inproceedings{xu20243d,
  title={3d gaussian parametric head model},
  author={Xu, Yuelang and Wang, Lizhen and Zheng, Zerong and Su, Zhaoqi and Liu, Yebin},
  booktitle={European Conference on Computer Vision},
  pages={129--147},
  year={2024},
  organization={Springer}
}

@inproceedings{dhamo2024headgas,
  title={Headgas: Real-time animatable head avatars via 3d gaussian splatting},
  author={Dhamo, Helisa and Nie, Yinyu and Moreau, Arthur and Song, Jifei and Shaw, Richard and Zhou, Yiren and P{\'e}rez-Pellitero, Eduardo},
  booktitle={European Conference on Computer Vision},
  pages={459--476},
  year={2024},
  organization={Springer}
}

@inproceedings{qian20243dgs,
  title={3dgs-avatar: Animatable avatars via deformable 3d gaussian splatting},
  author={Qian, Zhiyin and Wang, Shaofei and Mihajlovic, Marko and Geiger, Andreas and Tang, Siyu},
  booktitle={Proceedings of the IEEE/CVF conference on computer vision and pattern recognition},
  pages={5020--5030},
  year={2024}
}

@inproceedings{wei2025graphavatar,
  title={Graphavatar: Compact head avatars with gnn-generated 3d gaussians},
  author={Wei, Xiaobao and Chen, Peng and Lu, Ming and Chen, Hui and Tian, Feng},
  booktitle={Proceedings of the AAAI Conference on Artificial Intelligence},
  volume={39},
  pages={8295--8303},
  year={2025}
}

@inproceedings{ki2024learning,
  title={Learning to generate conditional tri-plane for 3d-aware expression controllable portrait animation},
  author={Ki, Taekyung and Min, Dongchan and Chae, Gyeongsu},
  booktitle={European Conference on Computer Vision},
  pages={476--493},
  year={2024},
  organization={Springer}
}

@article{li2023generalizable,
  title={Generalizable one-shot 3d neural head avatar},
  author={Li, Xueting and De Mello, Shalini and Liu, Sifei and Nagano, Koki and Iqbal, Umar and Kautz, Jan},
  journal={Advances in Neural Information Processing Systems},
  volume={36},
  pages={47239--47250},
  year={2023}
}

@inproceedings{ma2023otavatar,
  title={Otavatar: One-shot talking face avatar with controllable tri-plane rendering},
  author={Ma, Zhiyuan and Zhu, Xiangyu and Qi, Guo-Jun and Lei, Zhen and Zhang, Lei},
  booktitle={Proceedings of the IEEE/CVF Conference on Computer Vision and Pattern Recognition},
  pages={16901--16910},
  year={2023}
}

@inproceedings{tran2024voodoo,
  title={Voodoo 3d: Volumetric portrait disentanglement for one-shot 3d head reenactment},
  author={Tran, Phong and Zakharov, Egor and Ho, Long-Nhat and Tran, Anh Tuan and Hu, Liwen and Li, Hao},
  booktitle={Proceedings of the IEEE/CVF Conference on Computer Vision and Pattern Recognition},
  pages={10336--10348},
  year={2024}
}

@article{trevithick2023real,
  title={Real-time radiance fields for single-image portrait view synthesis},
  author={Trevithick, Alex and Chan, Matthew and Stengel, Michael and Chan, Eric and Liu, Chao and Yu, Zhiding and Khamis, Sameh and Chandraker, Manmohan and Ramamoorthi, Ravi and Nagano, Koki},
  journal={ACM Transactions on Graphics (TOG)},
  volume={42},
  number={4},
  pages={1--15},
  year={2023},
  publisher={ACM New York, NY, USA}
}

@inproceedings{yang2020facescape,
  title={Facescape: a large-scale high quality 3d face dataset and detailed riggable 3d face prediction},
  author={Yang, Haotian and Zhu, Hao and Wang, Yanru and Huang, Mingkai and Shen, Qiu and Yang, Ruigang and Cao, Xun},
  booktitle={Proceedings of the ieee/cvf conference on computer vision and pattern recognition},
  pages={601--610},
  year={2020}
}

@inproceedings{zielonka2022towards,
  title={Towards metrical reconstruction of human faces},
  author={Zielonka, Wojciech and Bolkart, Timo and Thies, Justus},
  booktitle={European conference on computer vision},
  pages={250--269},
  year={2022},
  organization={Springer}
}

@incollection{buehler2024cafca,
    title={Cafca: High-quality Novel View Synthesis of Expressive Faces from Casual Few-shot Captures},
    author={Marcel C. Buehler and Gengyan Li and Erroll Wood and Leonhard Helminger and Xu Chen and Tanmay Shah and Daoye Wang and Stephan Garbin and Sergio Orts-Escolano and Otmar Hilliges and Dmitry Lagun and Jérémy Riviere and Paulo Gotardo and Thabo Beeler and Abhimitra Meka and Kripasindhu Sarkar},
    year={2024},
    booktitle={ACM SIGGRAPH Asia 2024 Conference Paper},
    doi={10.1145/3680528.3687580},
    url={https://doi.org/10.1145/3680528}
}

@inproceedings{zheng2023pointavatar,
  title={Pointavatar: Deformable point-based head avatars from videos},
  author={Zheng, Yufeng and Yifan, Wang and Wetzstein, Gordon and Black, Michael J and Hilliges, Otmar},
  booktitle={Proceedings of the IEEE/CVF conference on computer vision and pattern recognition},
  pages={21057--21067},
  year={2023}
}

@inproceedings{li2023one,
  title={One-shot high-fidelity talking-head synthesis with deformable neural radiance field},
  author={Li, Weichuang and Zhang, Longhao and Wang, Dong and Zhao, Bin and Wang, Zhigang and Chen, Mulin and Zhang, Bang and Wang, Zhongjian and Bo, Liefeng and Li, Xuelong},
  booktitle={Proceedings of the IEEE/CVF Conference on Computer Vision and Pattern Recognition},
  pages={17969--17978},
  year={2023}
}

@inproceedings{ma2024cvthead,
  title={Cvthead: One-shot controllable head avatar with vertex-feature transformer},
  author={Ma, Haoyu and Zhang, Tong and Sun, Shanlin and Yan, Xiangyi and Han, Kun and Xie, Xiaohui},
  booktitle={Proceedings of the IEEE/CVF Winter Conference on Applications of Computer Vision},
  pages={6131--6141},
  year={2024}
}

@inproceedings{yang2024learning,
  title={Learning dense correspondence for nerf-based face reenactment},
  author={Yang, Songlin and Wang, Wei and Lan, Yushi and Fan, Xiangyu and Peng, Bo and Yang, Lei and Dong, Jing},
  booktitle={Proceedings of the AAAI Conference on Artificial Intelligence},
  volume={38},
  pages={6522--6530},
  year={2024}
}

@article{chu2024gpavatar,
  title={GPAvatar: Generalizable and precise head avatar from image (s)},
  author={Chu, Xuangeng and Li, Yu and Zeng, Ailing and Yang, Tianyu and Lin, Lijian and Liu, Yunfei and Harada, Tatsuya},
  journal={arXiv preprint arXiv:2401.10215},
  year={2024}
}

@article{ye2024real3d,
  title={Real3d-portrait: One-shot realistic 3d talking portrait synthesis},
  author={Ye, Zhenhui and Zhong, Tianyun and Ren, Yi and Yang, Jiaqi and Li, Weichuang and Huang, Jiawei and Jiang, Ziyue and He, Jinzheng and Huang, Rongjie and Liu, Jinglin and others},
  journal={arXiv preprint arXiv:2401.08503},
  year={2024}
}

@InProceedings{Wang_2025_CVPR,
    author    = {Wang, Shengze and Li, Xueting and Liu, Chao and Chan, Matthew and Stengel, Michael and Fuchs, Henry and De Mello, Shalini and Nagano, Koki},
    title     = {Coherent 3D Portrait Video Reconstruction via Triplane Fusion},
    booktitle = {Proceedings of the IEEE/CVF Conference on Computer Vision and Pattern Recognition (CVPR)},
    month     = {June},
    year      = {2025},
    pages     = {10712-10722}
}

@InProceedings{Yan_2025_WACV,
      author    = {Yan, Peizhi and Ward, Rabab and Tang, Qiang and Du, Shan},
      title     = {Gaussian Deja-vu: Creating Controllable 3D Gaussian Head-Avatars with Enhanced Generalization and Personalization Abilities},
      booktitle = {Proceedings of the Winter Conference on Applications of Computer Vision (WACV)},
      month     = {February},
      year      = {2025},
      pages     = {276-286}
  }

@inproceedings{li2024talkinggaussian,
  title={Talkinggaussian: Structure-persistent 3d talking head synthesis via gaussian splatting},
  author={Li, Jiahe and Zhang, Jiawei and Bai, Xiao and Zheng, Jin and Ning, Xin and Zhou, Jun and Gu, Lin},
  booktitle={European Conference on Computer Vision},
  pages={127--145},
  year={2024},
  organization={Springer}
}

@inproceedings{hu2024gaussianavatar,
  title={Gaussianavatar: Towards realistic human avatar modeling from a single video via animatable 3d gaussians},
  author={Hu, Liangxiao and Zhang, Hongwen and Zhang, Yuxiang and Zhou, Boyao and Liu, Boning and Zhang, Shengping and Nie, Liqiang},
  booktitle={Proceedings of the IEEE/CVF conference on computer vision and pattern recognition},
  pages={634--644},
  year={2024}
}

@inproceedings{wang2025mega,
  title={Mega: Hybrid mesh-gaussian head avatar for high-fidelity rendering and head editing},
  author={Wang, Cong and Kang, Di and Sun, Heyi and Qian, Shenhan and Wang, Zixuan and Bao, Linchao and Zhang, Song-Hai},
  booktitle={Proceedings of the Computer Vision and Pattern Recognition Conference},
  pages={26274--26284},
  year={2025}
}

@inproceedings{ma20243d,
  title={3d gaussian blendshapes for head avatar animation},
  author={Ma, Shengjie and Weng, Yanlin and Shao, Tianjia and Zhou, Kun},
  booktitle={ACM SIGGRAPH 2024 Conference Papers},
  pages={1--10},
  year={2024}
}

@inproceedings{liang2023benchmarking,
  title={Benchmarking algorithmic bias in face recognition: An experimental approach using synthetic faces and human evaluation},
  author={Liang, Hao and Perona, Pietro and Balakrishnan, Guha},
  booktitle={Proceedings of the IEEE/CVF International Conference on Computer Vision},
  pages={4977--4987},
  year={2023}
}

@inproceedings{giebenhain2024npga,
 author    = {Simon Giebenhain and Tobias Kirschstein and  Martin R{\"{u}}nz and Lourdes Agapito and Matthias Nie{\ss}ner},
 title     = {NPGA: Neural Parametric Gaussian Avatars},
 booktitle = {SIGGRAPH Asia 2024 Conference Papers (SA Conference Papers '24), December 3-6, Tokyo, Japan},
 doi       = {10.1145/3680528.3687689},
 isbn      = {979-8-4007-1131-2/24/12},
 year      = {2024},
}

@article{kingma2013auto,
  title={Auto-encoding variational bayes},
  author={Kingma, Diederik P and Welling, Max},
  journal={arXiv preprint arXiv:1312.6114},
  year={2013}
}

@inproceedings{deng2022gram,
  title={Gram: Generative radiance manifolds for 3d-aware image generation},
  author={Deng, Yu and Yang, Jiaolong and Xiang, Jianfeng and Tong, Xin},
  booktitle={Proceedings of the IEEE/CVF conference on computer vision and pattern recognition},
  pages={10673--10683},
  year={2022}
}

@article{karras2021alias,
  title={Alias-free generative adversarial networks},
  author={Karras, Tero and Aittala, Miika and Laine, Samuli and H{\"a}rk{\"o}nen, Erik and Hellsten, Janne and Lehtinen, Jaakko and Aila, Timo},
  journal={Advances in neural information processing systems},
  volume={34},
  pages={852--863},
  year={2021}
}

@article{dhariwal2021diffusion,
  title={Diffusion models beat gans on image synthesis},
  author={Dhariwal, Prafulla and Nichol, Alexander},
  journal={Advances in neural information processing systems},
  volume={34},
  pages={8780--8794},
  year={2021}
}

@article{razavi2019generating,
  title={Generating diverse high-fidelity images with vq-vae-2},
  author={Razavi, Ali and Van den Oord, Aaron and Vinyals, Oriol},
  journal={Advances in neural information processing systems},
  volume={32},
  year={2019}
}

@inproceedings{li2024spherehead,
  title={Spherehead: stable 3d full-head synthesis with spherical tri-plane representation},
  author={Li, Heyuan and Chen, Ce and Shi, Tianhao and Qiu, Yuda and An, Sizhe and Chen, Guanying and Han, Xiaoguang},
  booktitle={European Conference on Computer Vision},
  pages={324--341},
  year={2024},
  organization={Springer}
}

@article{sun2022ide,
  title={Ide-3d: Interactive disentangled editing for high-resolution 3d-aware portrait synthesis},
  author={Sun, Jingxiang and Wang, Xuan and Shi, Yichun and Wang, Lizhen and Wang, Jue and Liu, Yebin},
  journal={ACM Transactions on Graphics (ToG)},
  volume={41},
  number={6},
  pages={1--10},
  year={2022},
  publisher={ACM New York, NY, USA}
}

@inproceedings{cao2024dreamavatar,
  title={Dreamavatar: Text-and-shape guided 3d human avatar generation via diffusion models},
  author={Cao, Yukang and Cao, Yan-Pei and Han, Kai and Shan, Ying and Wong, Kwan-Yee K},
  booktitle={Proceedings of the IEEE/CVF conference on computer vision and pattern recognition},
  pages={958--968},
  year={2024}
}

@inproceedings{zhou2024headstudio,
  title={Headstudio: Text to animatable head avatars with 3d gaussian splatting},
  author={Zhou, Zhenglin and Ma, Fan and Fan, Hehe and Yang, Zongxin and Yang, Yi},
  booktitle={European Conference on Computer Vision},
  pages={145--163},
  year={2024},
  organization={Springer}
}

@inproceedings{liang2024luciddreamer,
  title={Luciddreamer: Towards high-fidelity text-to-3d generation via interval score matching},
  author={Liang, Yixun and Yang, Xin and Lin, Jiantao and Li, Haodong and Xu, Xiaogang and Chen, Yingcong},
  booktitle={Proceedings of the IEEE/CVF conference on computer vision and pattern recognition},
  pages={6517--6526},
  year={2024}
}

@article{wang2023prolificdreamer,
  title={Prolificdreamer: High-fidelity and diverse text-to-3d generation with variational score distillation},
  author={Wang, Zhengyi and Lu, Cheng and Wang, Yikai and Bao, Fan and Li, Chongxuan and Su, Hang and Zhu, Jun},
  journal={Advances in neural information processing systems},
  volume={36},
  pages={8406--8441},
  year={2023}
}

@inproceedings{chen2023fantasia3d,
  title={Fantasia3d: Disentangling geometry and appearance for high-quality text-to-3d content creation},
  author={Chen, Rui and Chen, Yongwei and Jiao, Ningxin and Jia, Kui},
  booktitle={Proceedings of the IEEE/CVF international conference on computer vision},
  pages={22246--22256},
  year={2023}
}

@incollection{liu2024stylegaussian,
  title={Stylegaussian: Instant 3d style transfer with gaussian splatting},
  author={Liu, Kunhao and Zhan, Fangneng and Xu, Muyu and Theobalt, Christian and Shao, Ling and Lu, Shijian},
  booktitle={SIGGRAPH Asia 2024 Technical Communications},
  pages={1--4},
  year={2024}
}

@inproceedings{li2024photomaker,
  title={Photomaker: Customizing realistic human photos via stacked id embedding},
  author={Li, Zhen and Cao, Mingdeng and Wang, Xintao and Qi, Zhongang and Cheng, Ming-Ming and Shan, Ying},
  booktitle={Proceedings of the IEEE/CVF conference on computer vision and pattern recognition},
  pages={8640--8650},
  year={2024}
}

@article{li2023blip,
  title={Blip-diffusion: Pre-trained subject representation for controllable text-to-image generation and editing},
  author={Li, Dongxu and Li, Junnan and Hoi, Steven},
  journal={Advances in Neural Information Processing Systems},
  volume={36},
  pages={30146--30166},
  year={2023}
}

@inproceedings{wang2023exploring,
  title={Exploring clip for assessing the look and feel of images},
  author={Wang, Jianyi and Chan, Kelvin CK and Loy, Chen Change},
  booktitle={Proceedings of the AAAI conference on artificial intelligence},
  volume={37},
  number={2},
  pages={2555--2563},
  year={2023}
}

@article{heusel2017gans,
  title={Gans trained by a two time-scale update rule converge to a local nash equilibrium},
  author={Heusel, Martin and Ramsauer, Hubert and Unterthiner, Thomas and Nessler, Bernhard and Hochreiter, Sepp},
  journal={Advances in neural information processing systems},
  volume={30},
  year={2017}
}

@inproceedings{deng2020retinaface,
  title={Retinaface: Single-shot multi-level face localisation in the wild},
  author={Deng, Jiankang and Guo, Jia and Ververas, Evangelos and Kotsia, Irene and Zafeiriou, Stefanos},
  booktitle={Proceedings of the IEEE/CVF conference on computer vision and pattern recognition},
  pages={5203--5212},
  year={2020}
}

@article{feng2021deca,
author = {Feng, Yao and Feng, Haiwen and Black, Michael J. and Bolkart, Timo},
title = {Learning an animatable detailed 3D face model from in-the-wild images},
year = {2021},
issue_date = {August 2021},
publisher = {Association for Computing Machinery},
address = {New York, NY, USA},
volume = {40},
number = {4},
issn = {0730-0301},
url = {https://doi.org/10.1145/3450626.3459936},
doi = {10.1145/3450626.3459936},
journal = {ACM Trans. Graph.},
month = jul,
articleno = {88},
numpages = {13}
}

}




\newpage
\appendix

\section{Technical Appendices and Supplementary Material}

\subsection{Implementation Details}
\label{sec:impl_details}

\subsubsection{Pretrained Models}
We use Stable Diffusion v1.5 (Realistic Vision V4.0) with the stabilityai/sd-vae-ft-mse VAE as the diffusion backbone, running in float16 precision. Identity conditioning is provided by IP-Adapter Plus Face using a CLIP ViT-H/14 image encoder (scale 0.8), and pose conditioning by ControlNet (control\_v11p\_sd15\_openpose, scale 0.6). For 3DGS, we use gsplat with base models pretrained on NeRSemble~\cite{kirschstein2023nersemble} sequences. Each base model uses degree-3 spherical harmonics and contains approximately 100K Gaussians, stored in PLY format. All renderings use a black background.

\subsubsection{Base 3DGS Model Selection}
\label{sec:hair_alg}
Given an input identity image, we select the most geometrically compatible NeRSemble base 3DGS by computing weighted feature similarity in the embedding space of DINO ViT-B/16~\cite{caron2021emerging}. Specifically, we extract patch-level features from three semantically meaningful facial regions (hair, skin, and overall face shape) and compute a weighted cosine similarity score with weights $0.4$, $0.3$, and $0.3$ respectively. These weights were chosen to prioritize hair structure, which most strongly determines the geometric complexity of the 3D reconstruction, while also accounting for skin tone and overall facial geometry.

\subsubsection{Diffusion Denoising}
We use a DDIM scheduler with $T = 1000$ timesteps and a linear beta schedule ($\beta_\mathrm{start} = 0.00085$, $\beta_\mathrm{end} = 0.012$). We run 50 inference steps with denoising strength $s = 0.6$ ($t_s = 600$), classifier-free guidance scale 4.0, and a null text input. The hybrid clean prediction weight is fixed at $w = 0.4$ throughout denoising; we do not anneal $w$ because at later steps the latent $\mathbf{x}_t$ dominates the prediction and the noise component has diminishing influence.

\subsubsection{3DGS Refitting}
At each guidance step, we refit the 3DGS using separate Adam optimizers per parameter group (positions: $1.6 \times 10^{-4}$, scales: $5 \times 10^{-3}$, rotations: $1 \times 10^{-3}$, opacities: $5 \times 10^{-2}$, SH$_0$: $2.5 \times 10^{-3}$, SH$_{1+}$: $1.25 \times 10^{-4}$; all with $\epsilon = 10^{-15}$). The first guidance step uses $M = 1{,}000$ refit iterations; subsequent steps use $M = 200$. The photometric loss uses $\lambda_1 = 0.8$ (L1) and $\lambda_2 = 0.2$ (SSIM), sampling one random view per iteration. The geometry guidance weight is $\lambda = 50{,}000$, with a loss scaling factor of 1000 for numerical stability during backpropagation through the VAE decoder. We do not perform Gaussian densification during refitting, as it causes tensor size mismatches when parameters are carried across timesteps.

\subsubsection{Computational Cost}
We report end-to-end runtime per identity under our evaluation setting, including preprocessing, generation or optimization, and rendering of evaluation views when required.
Our full pipeline takes approximately 3 minutes per identity on a single NVIDIA A100 GPU, with peak memory usage of approximately 24GB.
For comparison, FastAvatar~\cite{liang2025fastavatar} takes about 3 seconds, FaceLift~\cite{lyu2025facelift} about 20 seconds, LAM~\cite{he2025lam} about 65 seconds, GAGAvatar~\cite{chu2024generalizable} about 107 seconds, DreamGaussian~\cite{tang2023dreamgaussian} about 182 seconds, GaussianEdit~\cite{chen2024gaussianeditor,wang2024view,wen2025intergsedit} about 4 minutes, Human-3Diffusion~\cite{xue2024human} about 5 minutes, and Arc2Avatar~\cite{gerogiannis2025arc2avatar} about 3.5 hours per identity.
Thus, our method is slower than feed-forward avatar methods, but comparable to or faster than optimization-heavy and diffusion-based 3D generation baselines, while producing an explicit 3DGS avatar as the final output.
For the quantitative evaluation on 100 identities, our method requires roughly 5 A100 GPU-hours; scaling to the full 6,279-image evaluation would require approximately 314 A100 GPU-hours.

\subsubsection{Evaluation Protocol}
We evaluate all methods using a unified manifest-based protocol. 
For each method, generated outputs are organized by identity and paired with the corresponding source image. 
Generated views are sorted by parsed camera angle when available, and otherwise by natural filename order. 
We use all generated views for cross-view consistency and no-reference perceptual quality, and use the six most frontal views for identity similarity and FID. 
Before metric computation, images are padded to square and resized to avoid cropping off-center faces.

For methods that produce an explicit 3D representation, we render the final representation from the evaluation cameras. 
For methods whose released pipelines directly output generated or rendered views, we evaluate those outputs under the closest matching camera or pose setting. 
For \alg{}, all reported metrics are computed on the final 3DGS renderings rather than intermediate diffusion predictions, since the explicit 3DGS avatar is the final output.

\begin{figure}[h]
    \centering
    \includegraphics[height=0.9\textheight]{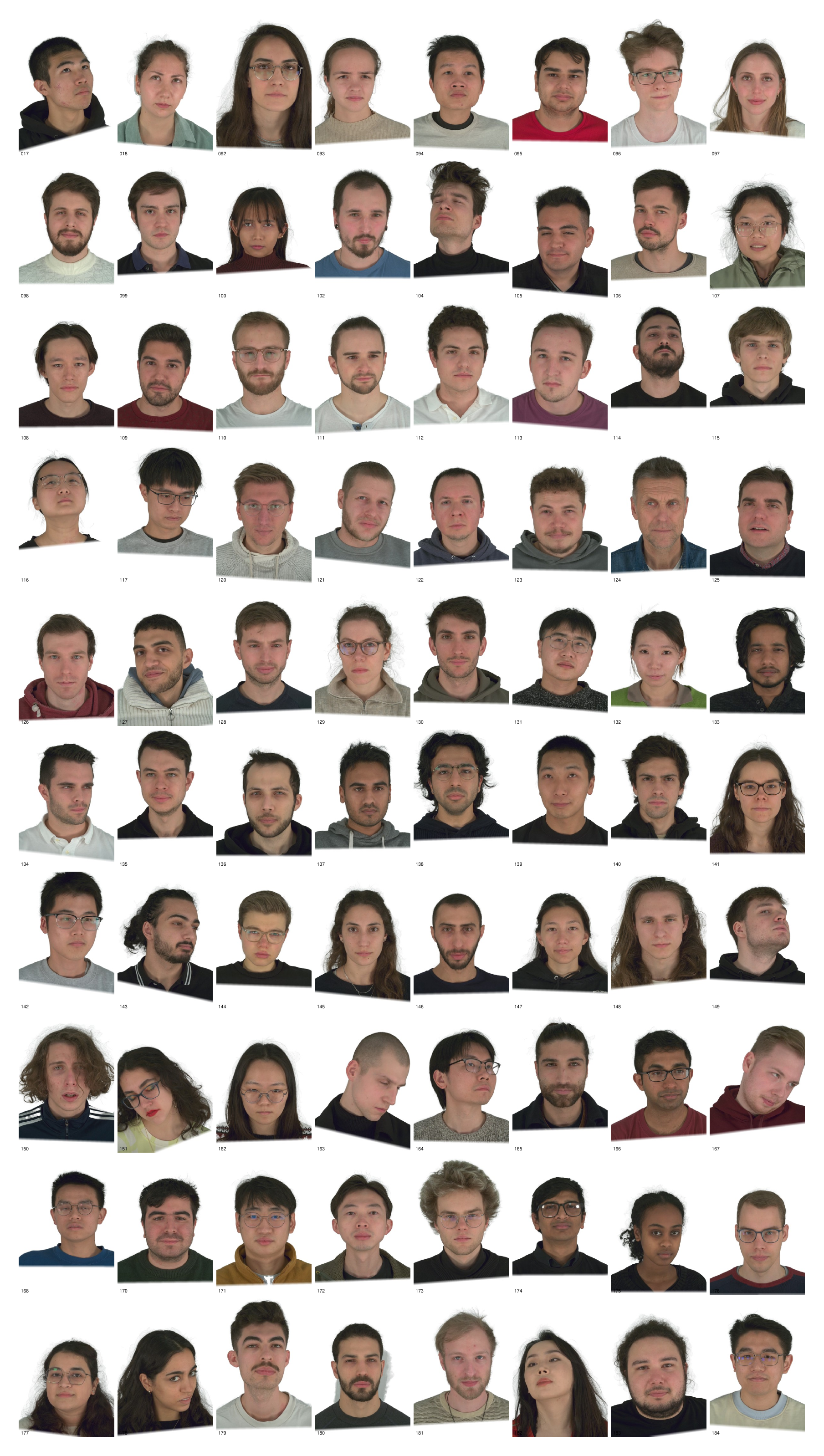}
    \caption{Example identities in the NeRSemble dataset. These sequences serve as the geometric base for our 3D avatar generation pipeline.}
    \label{fig:nersemble_bases}
\end{figure}

\subsection{Algorithm of \alg{}}
Algorithm~\ref{alg:main} summarizes the complete geometry-guided denoising procedure described in \S\ref{sec:method}.

\begin{algorithm}[H]
\caption{3DGS-Guided Iterative Denoising}
\label{alg:main}
\begin{algorithmic}[1]
\Require Base 3DGS model $\mathcal{M}$, input image $I_\mathrm{in}$, $V$ camera views
\Require Diffusion model $(\mathcal{E}, \mathcal{D}, \boldsymbol{\epsilon}_\theta)$, guidance weight $\lambda$, noise mixture weight $w$, refit steps $N$, strength $s$
\Ensure Final 3DGS model $\mathcal{M}$, generated views $\{\hat{I}^v\}$
\State Render base views: $I^v \gets \mathcal{R}(\mathcal{M}, v)$ for all $v$
\State Encode conditioning: $\mathbf{c} \gets \text{ImageAdapter}(I_\mathrm{in})$
\State Encode to latent: $\mathbf{x}_0^v \gets \mathcal{E}(I^v)$ for all $v$
\State Sample noise: $\boldsymbol{\epsilon}_\mathrm{gt}^v \sim \mathcal{N}(\mathbf{0}, \mathbf{I})$ for all $v$
\State Compute starting step: $t_s \gets \lfloor s \cdot T \rfloor$
\State Add noise: $\mathbf{x}_{t_s}^v \gets \alpha_{t_s} \mathbf{x}_0^v + \sigma_{t_s} \boldsymbol{\epsilon}_\mathrm{gt}^v$ \Comment{Eq.~\ref{eq:forward}}
\For{$t = t_s, \ldots, 1$}
    \For{$v = 1, \ldots, V$} \Comment{\textbf{Step 1}: Per-view generation}
        \State $\boldsymbol{\epsilon}_\theta^{t,v} \gets \boldsymbol{\epsilon}_\theta(\mathbf{x}_t^v, t, \mathbf{c})$
        \State $\boldsymbol{\epsilon}_\mathrm{mix}^{t,v} \gets w \cdot \boldsymbol{\epsilon}_\theta^{t,v} + (1 - w) \cdot \boldsymbol{\epsilon}_\mathrm{gt}^v$ \Comment{Eq.~\ref{eq:hybrid_noise}}
        \State $\hat{\mathbf{x}}_0^v \gets (\mathbf{x}_t^v - \sigma_t \boldsymbol{\epsilon}_\mathrm{mix}^{t,v}) / \alpha_t$ \Comment{Eq.~\ref{eq:ddim_x0}}
        \State $\hat{I}^v \gets \mathcal{D}(\hat{\mathbf{x}}_0^v)$
    \EndFor
    \For{$n = 1, \ldots, N$} \Comment{\textbf{Step 2}: 3DGS model update}
        \State Sample view $v$, render $I_r^v \gets \mathcal{R}(\mathcal{M}, v)$
        \State Update $\mathcal{M}$ by $\nabla_\mathcal{M} \mathcal{L}_\mathrm{photo}(I_r^v, \hat{I}^v)$ \Comment{Eq.~\ref{eq:photo}}
    \EndFor
    \For{$v = 1, \ldots, V$} \Comment{\textbf{Step 3}: Noise adjustment}
        \State $\mathbf{g} \gets \mathcal{L}_\mathrm{photo}(\hat{I}^v, \mathcal{R}(\mathcal{M}, v))$
        \State $\boldsymbol{\epsilon}_\theta^{t,v} \gets \boldsymbol{\epsilon}_\theta^{t,v} - \lambda \cdot \partial \mathbf{g} / \partial \boldsymbol{\epsilon}_\theta^{t,v}$ \Comment{Eq.~\ref{eq:noise_adjust}}
    \EndFor
    \For{$v = 1, \ldots, V$} \Comment{DDIM step}
        \State $\mathbf{x}_{t-1}^v \gets \alpha_{t-1} \hat{\mathbf{x}}_0^v + \sigma_{t-1} \boldsymbol{\epsilon}_\theta^{t,v}$ \Comment{Eq.~\ref{eq:ddim_step}}
    \EndFor
\EndFor
\State \Return $\mathcal{M}$, $\{\hat{I}^v\}$
\end{algorithmic}
\end{algorithm}

\subsection{3D Consistency of Diffusion-based Baseline Outputs}
\label{sec:novelview}
\begin{figure}[H]
    \centering
    \includegraphics[width=\columnwidth]{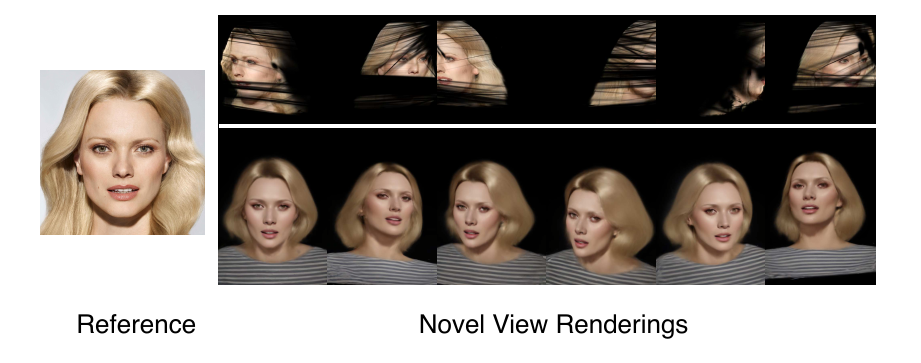}
    \caption{\textbf{Novel view renderings from refitted 3DGS.} Top: GAGAvatar's outputs refitted to a 3DGS and rendered from novel viewpoints. The severe artifacts (floaters, fragmented geometry, inconsistent structure) reveal that GAGAvatar's multi-view outputs are not 3D-consistent, as its visual quality relies on a 2D neural renderer rather than an explicit 3D representation. Bottom: our method's outputs refitted under the same procedure, producing clean and coherent novel views.}
    \label{fig:novel_view_refit}
\end{figure}

Methods that rely on a 2D enhancer or neural renderer to produce their final output may appear high-quality when viewed individually, but their outputs are not guaranteed to be 3D-consistent. To demonstrate this, we take GAGAvatar\cite{chu2024generalizable}'s multi-view outputs, fit a 3DGS model using the same cameras and optimization procedure, and render from novel viewpoints. As shown in Fig.~\ref{fig:novel_view_refit}, GAGAvatar's refitted 3DGS produces severe artifacts including floaters, fragmented geometry, and broken structure, confirming that the original outputs cannot be reconciled into a coherent 3D model. In contrast, our outputs produce clean novel-view renderings under the same procedure, as the geometry guidance during denoising ensures that the generated views are already mutually consistent in 3D.

\subsection{The effect of base 3DGS model}
\label{sec:supp_base}
\begin{figure}[t]
    \centering
    \includegraphics[width=.9\columnwidth]{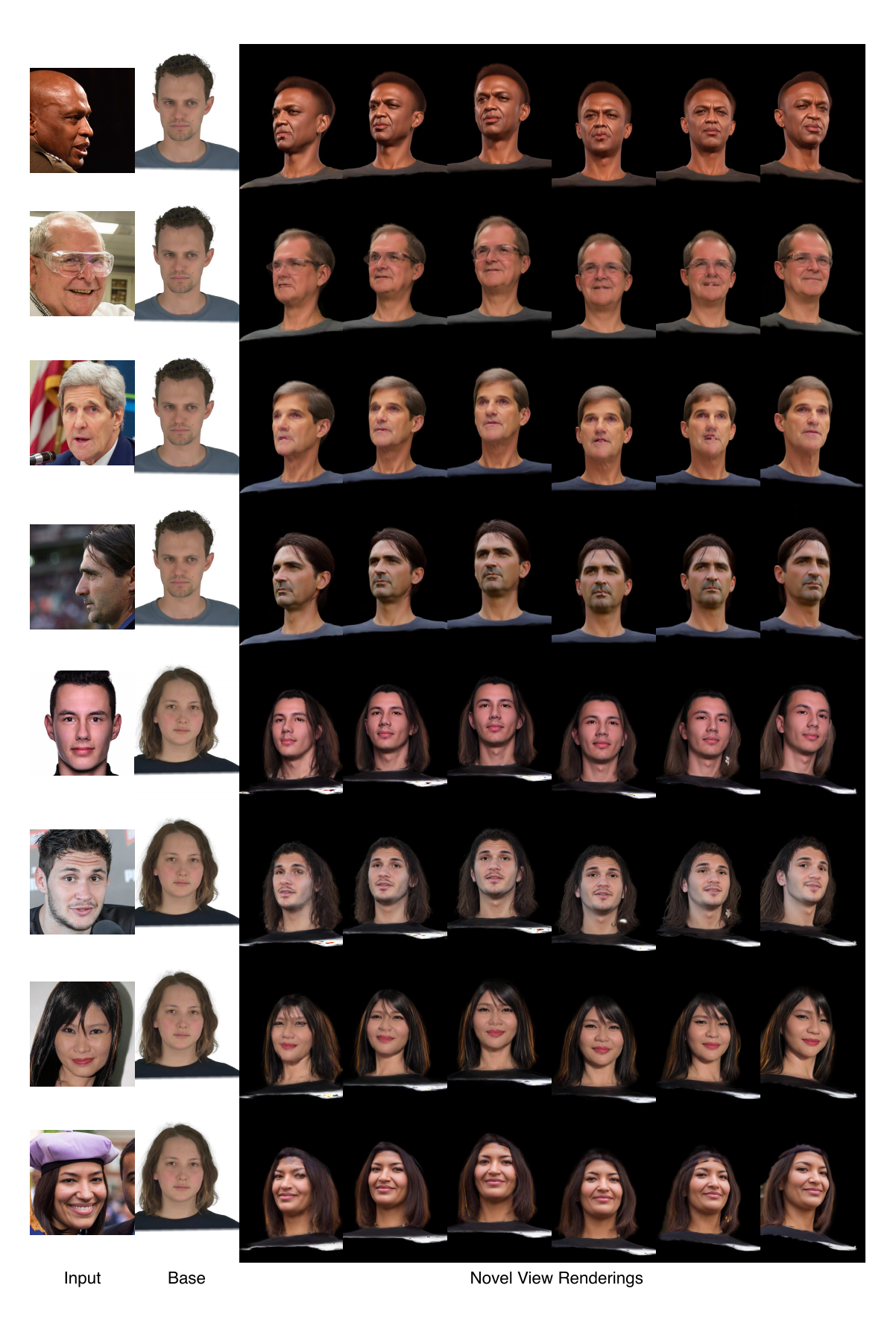}
    \caption{\textbf{Same base model, different input images.} Top 4 and bottom 4 models have different input images (left), and same base 3DGS model (second column). }
    \label{fig:same_base}
\end{figure}

\begin{figure}[t]
    \centering
    \includegraphics[width=.9\columnwidth]{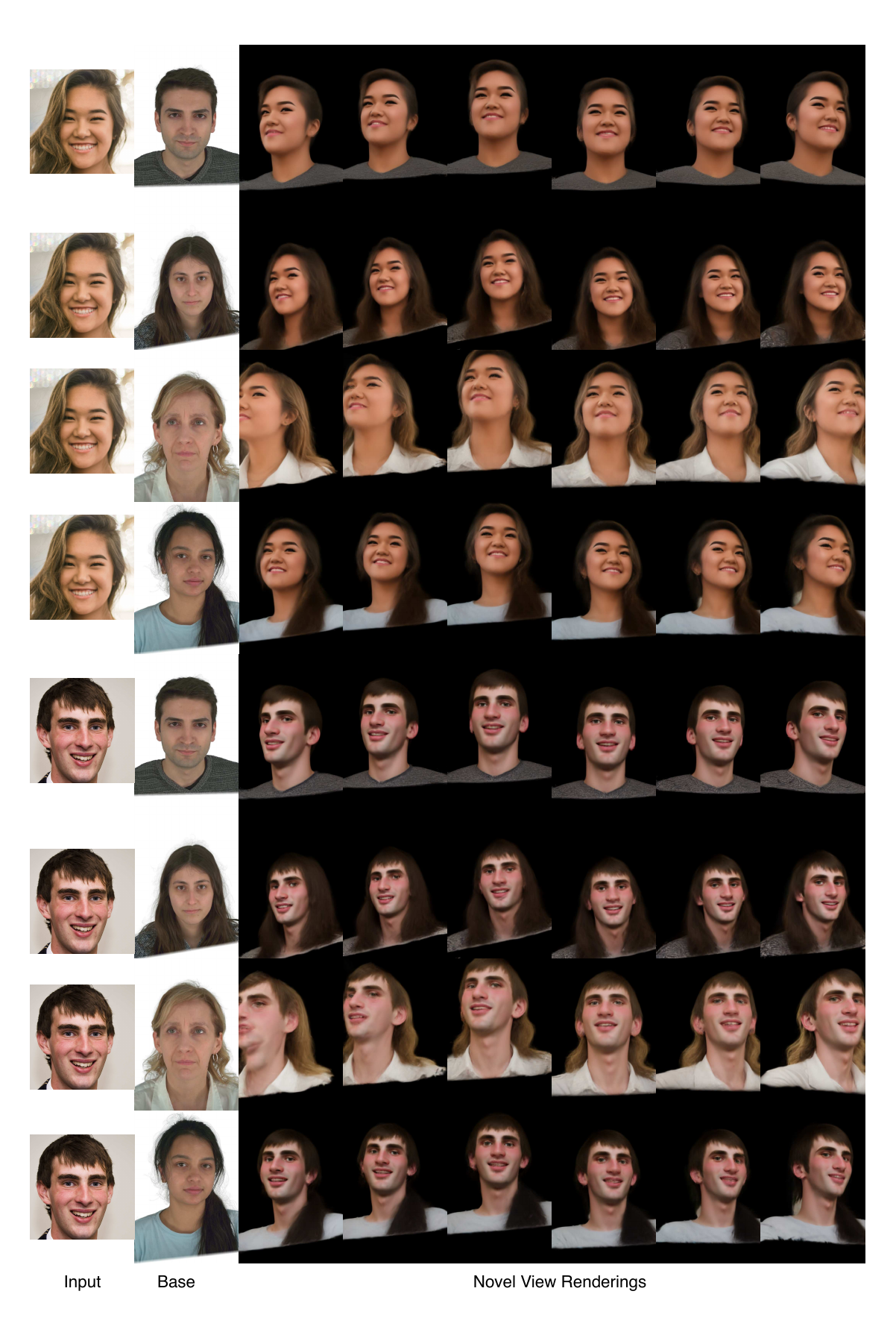}
    \caption{\textbf{Same input image, different base models.} Top 4 and bottom 4 models share a same input  image (left), with different 3DGS base models (second column). }
    \label{fig:same_ref}
\end{figure}

We investigate the role of the base 3DGS model through two complementary experiments.

\paragraph{Same base, different inputs (Fig.~\ref{fig:same_base}).} We fix a single base model and vary the input across diverse identities. In all cases, facial features (skin tone, eye shape, facial structure) change to match the input, while hairstyle remain largely inherited from the base. This holds consistently across both base models shown (top and bottom four rows use different bases). The pattern arises because trained identity encoders like IP-Adapter operate on tightly cropped face regions that exclude most hair, so hair is largely absent from the identity signal. Additionally, hair geometry is inherently less stable for 3DGS due to its fine, semi-transparent structure, making the model more resistant to hair modifications during refitting.

\paragraph{Same input, different bases (Fig.~\ref{fig:same_ref}).} We fix a single input and generate avatars using four different base models per identity. The reference identity is preserved across all bases, but hairstyle and clothing vary with each base. Notably, facial features such as nose shape and jawline do adapt to the input even when the base has a substantially different face structure, confirming that the geometry guidance acts as a soft anchor that permits structural changes where the identity signal is strong. This disentanglement motivates our base selection algorithm, which prioritizes hairstyle similarity over other attributes, since hairstyle is the feature most inherited from the base and least controllable through identity conditioning.

\subsection{Additional Qualitative Results}

\begin{figure}[t]
    \centering
    \includegraphics[width=\textwidth]{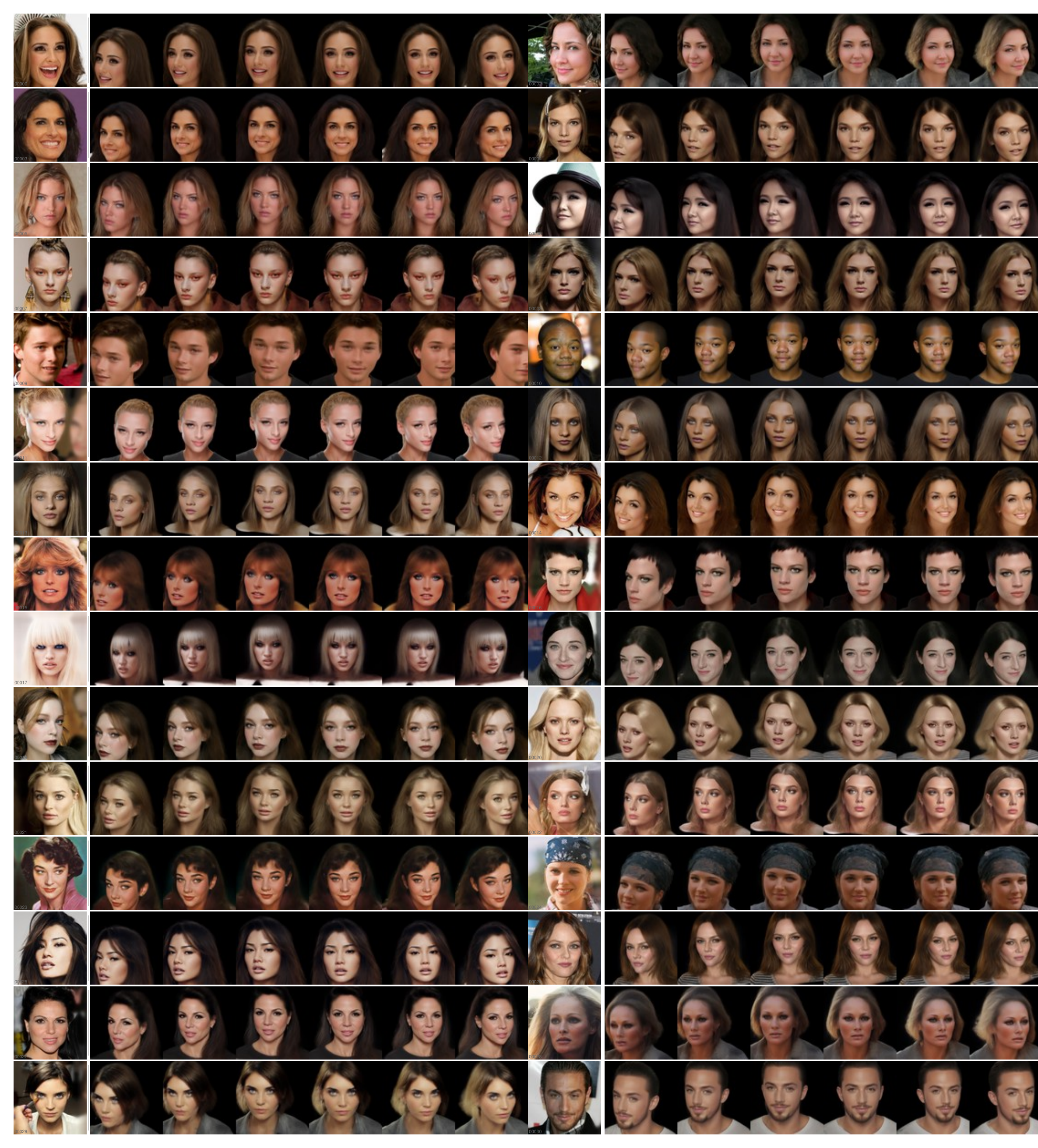}
    \caption{CelebA-3D generation results (identities 00000--00029). For each identity, the leftmost image is the input CelebA-HQ photograph, followed by six rendered views of the generated 3D Gaussian avatar at evenly spaced azimuth angles.}
    \label{fig:celeba3d_results_01}
\end{figure}

\begin{figure}[t]
    \centering
    \includegraphics[width=\textwidth]{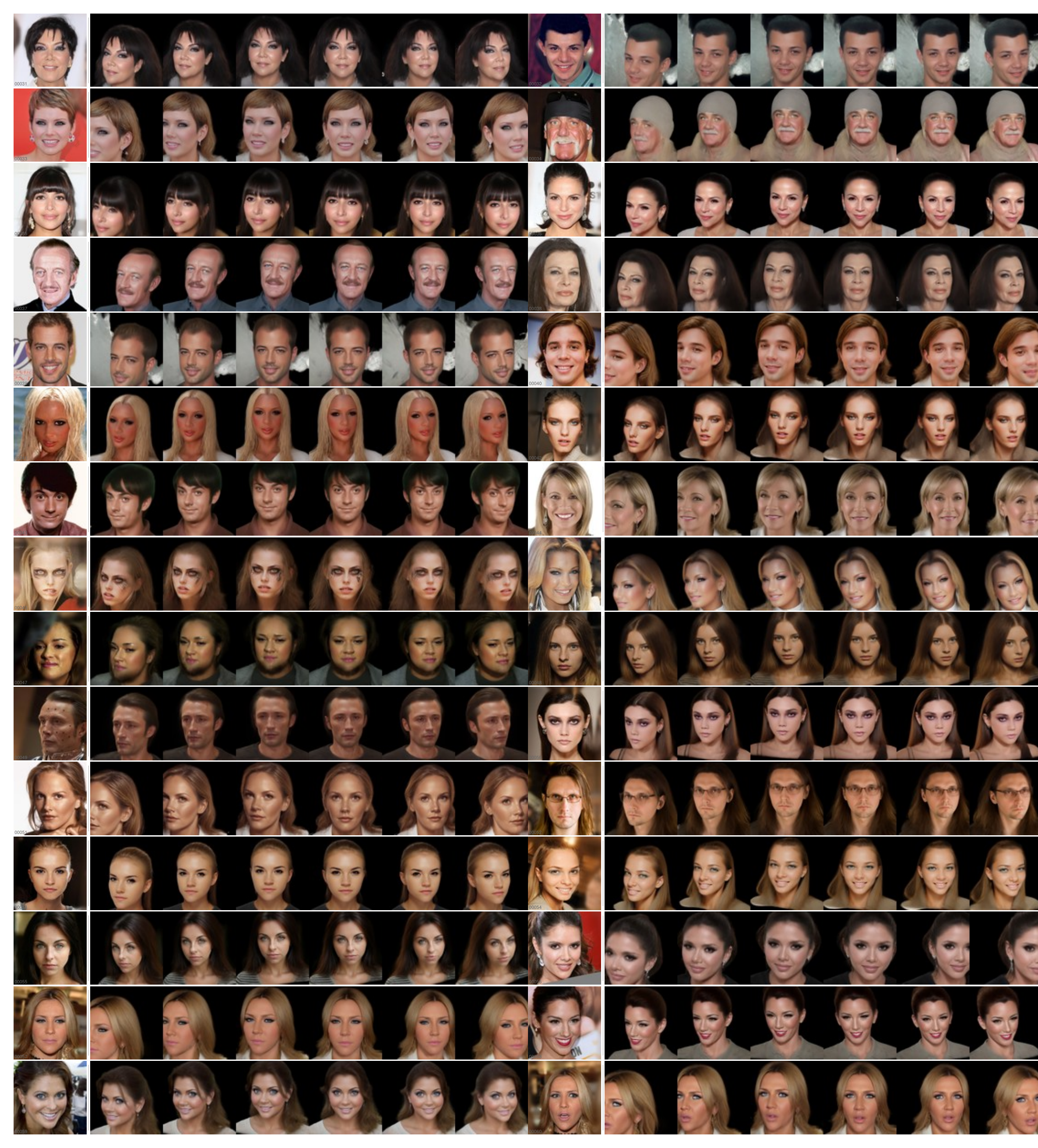}
    \caption{CelebA-3D generation results (identities 00030--00059).}
    \label{fig:celeba3d_results_02}
\end{figure}

\begin{figure}[t]
    \centering
    \includegraphics[width=\textwidth]{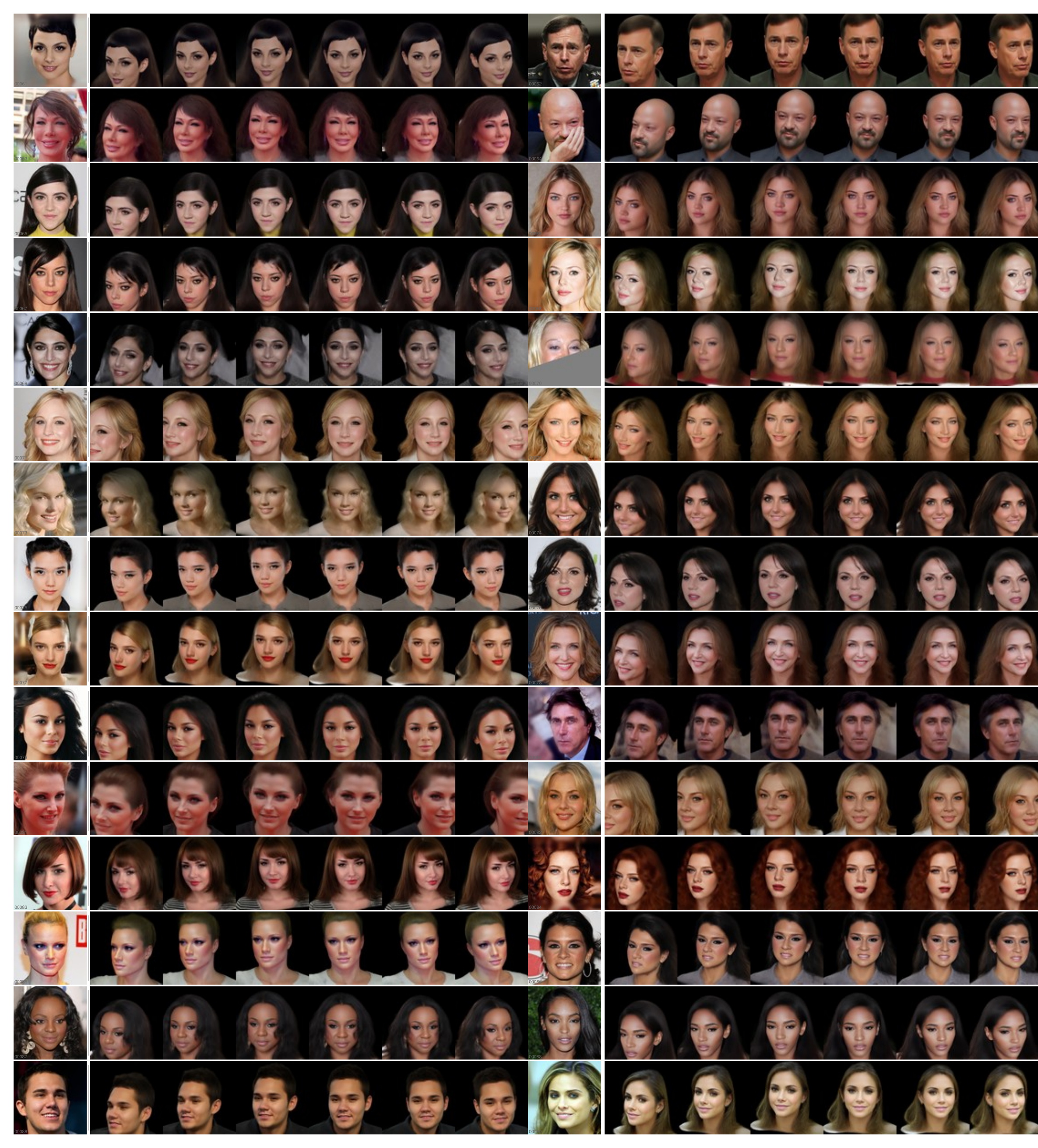}
    \caption{CelebA-3D generation results (identities 00060--00089).}
    \label{fig:celeba3d_results_03}
\end{figure}

We present qualitative results of our method on the CelebA-3D dataset in Figures~\ref{fig:celeba3d_results_01}, \ref{fig:celeba3d_results_02}, \ref{fig:celeba3d_results_03}. For each identity, we show the input CelebA-HQ photograph alongside six rendered views of the generated 3D Gaussian avatar at evenly spaced azimuth angles, demonstrating consistent geometry and appearance across viewpoints. Results span a diverse range of identities including varied ages, ethnicities, hair styles, and lighting conditions, highlighting the generalization capability of our pipeline.

\subsection{Multi-view Outputs from Diffusion v.s. 3DGS Renderings}

\begin{figure}[t]
    \centering
    \includegraphics[height=0.9\textheight]{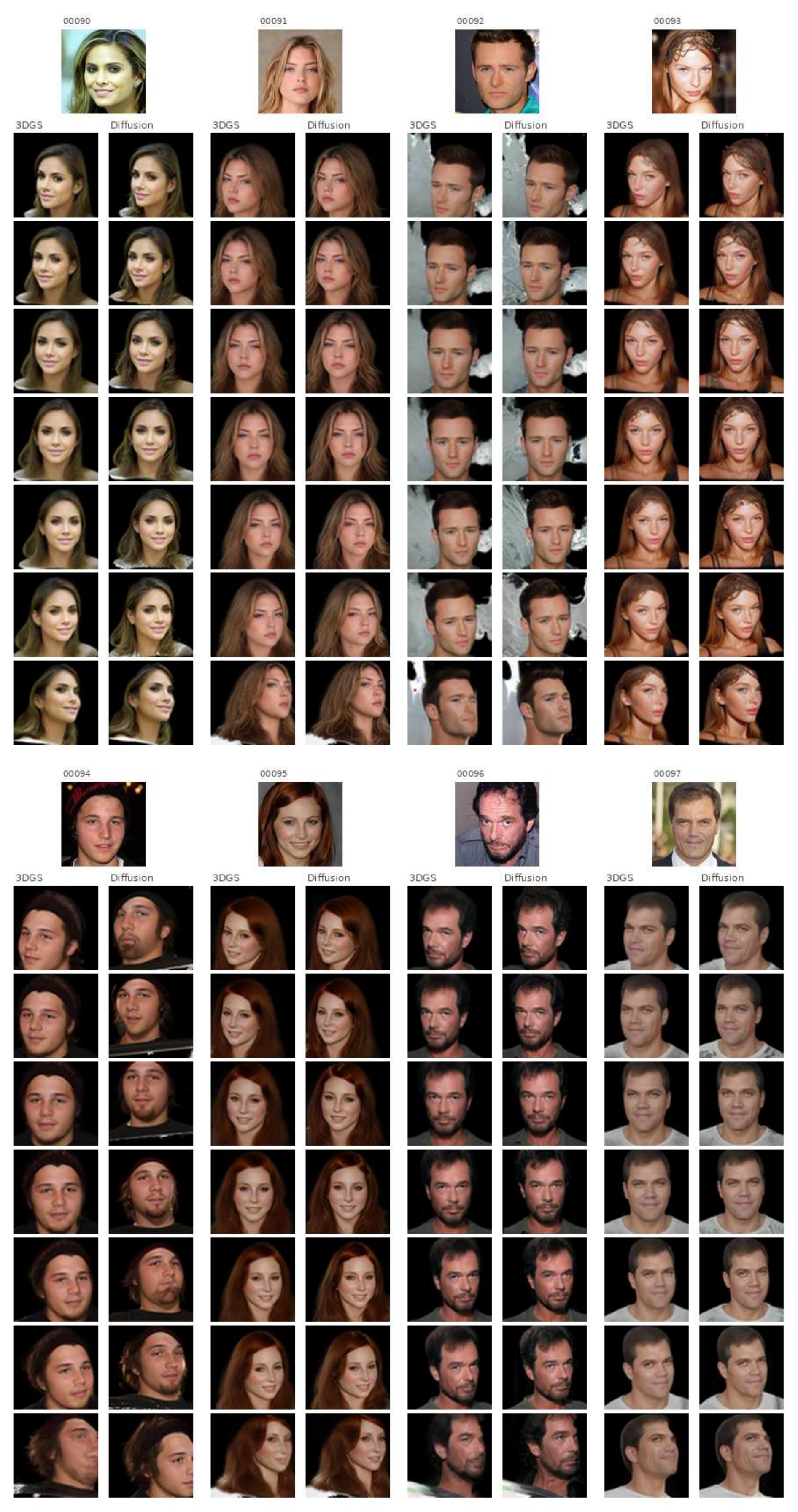}
    \caption{Qualitative comparison between diffusion outputs and 3DGS renderings for identities 00090--00097. For each identity, the source CelebA-HQ photograph is shown at the top, with the 3DGS-rendered views (left column) and diffusion-generated views (right column).}
    \label{fig:comparison_grid}
\end{figure}

Fig.~\ref{fig:comparison_grid} shows additional results on diverse CelebA-HQ identities. For each identity, we display the input image alongside multi-view renderings from both the final 3DGS model and the diffusion outputs. Across varying genders, ages, skin tones, and lighting conditions, the method consistently preserves the reference identity while maintaining cross-view geometric coherence. The 3DGS renderings are slightly smoother due to the Gaussian representation, while the diffusion outputs retain finer texture details.

\subsection{Alternative approaches and negative results}

\subsubsection{GaussianEdit: Attention-Based Baseline Implementation}
\label{sec:supp_gaussianedit}

\begin{figure}[H]
    \centering
    \includegraphics[width=0.8\textwidth]{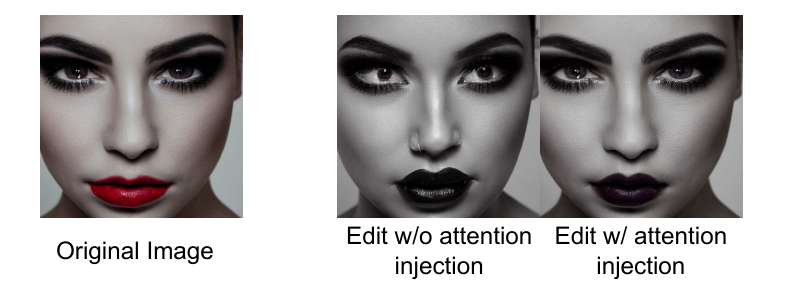}
    \caption{\textbf{Attention injection for text-guided editing.} Left: source image. Middle: editing with attention injection preserves structure and localizes the change. Right: editing without injection causes unintended structural changes.}
    \label{fig:attn_inject}
\end{figure}

\paragraph{Background: attention injection for text-guided editing.} Cross-attention maps in the diffusion UNet control where each conditioning token influences the generated image. For text-guided editing\cite{hertz2022prompt}, injecting attention maps from a source prompt into an edited prompt enables localized edits while preserving unrelated structure. Fig.~\ref{fig:attn_inject} illustrates this: changing ``red lipstick'' to ``dark lipstick'' without attention injection alters the entire face structure (right), while injecting the source cross-attention maps for the first 80\% of denoising timesteps restricts the edit to the lips only (middle), leaving other features intact. This demonstrates that text tokens have spatially disentangled control over generation.

\paragraph{Extension to 3DGS: inverse-forward rendering.} To enforce this consistency across multiple views of a 3DGS scene, prior work~\cite{chen2024gaussianeditor,wen2025intergsedit,wang2024view} inverse-renders the 2D attention maps onto the 3D Gaussians. For each Gaussian $G_k$, its attention value for token $n$ is computed by accumulating 2D attention across all views, weighted by alpha-compositing weights:
\begin{equation}
    a_k^{(n)} = \frac{\sum_{v} w_k^v \cdot \mathbf{A}^v(\pi_v(G_k), n)}{\sum_{v} w_k^v},
\end{equation}
where $\pi_v(G_k)$ is the projection of $G_k$ in view $v$ and $w_k^v$ is its compositing weight. These 3D-consistent values are then forward-rendered back to 2D via standard splatting:
\begin{equation}
    \tilde{\mathbf{A}}^v(p, n) = \sum_{k} a_k^{(n)} \cdot \alpha_k^v(p) \prod_{j < k} (1 - \alpha_j^v(p)),
\end{equation}
replacing the original per-view attention maps in the UNet so that all views share a geometrically consistent attention distribution.

\paragraph{Our reimplementation.} We apply this inverse-forward procedure to \emph{every} image token from IP-Adapter conditioning, rather than text tokens. At each denoising step, we extract cross-attention maps between spatial features and image tokens, perform the 3D aggregation and re-projection using $\mathcal{M}$ and known cameras, and substitute the original maps. As shown in Table~\ref{tab:quantitative} and Fig.~\ref{fig:main_results}, this yields limited improvement: unlike text tokens, image tokens activate broadly and lack spatial disentanglement (Fig.~\ref{fig:attention}), so the 3D-aggregated maps are themselves poorly localized, and the consistency enforcement blends incompatible view-dependent information rather than resolving it.

\subsection{PeRFlow: Accelerated Sampling via Rectified Flow}

To reduce runtime, we experimented with PeRFlow~\cite{yan2024perflow}, which is based on rectified flow matching~\cite{liu2022flow}. Unlike standard diffusion models that learn a gradual denoising process over many steps, flow matching learns a direct transport map between the noise distribution $\mathcal{N}(\mathbf{0}, \mathbf{I})$ and the data distribution along straight-line trajectories in latent space. PeRFlow distills these trajectories into a few-step process, enabling high-quality generation in as few as 4-8 function evaluations compared to the 50 steps used in our standard DDIM pipeline.

\begin{figure}[H]
    \centering
    \includegraphics[width=\textwidth]{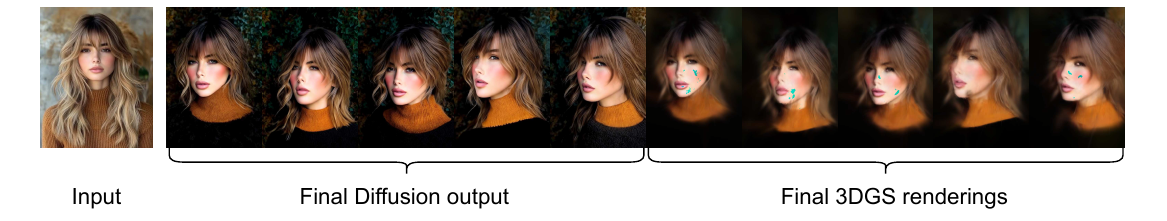}
    \caption{\textbf{PeRFlow results.} With only 4 sampling steps, the diffusion outputs (left) appear reasonable individually but the 3DGS renderings (right) reveal poor multi-view consistency due to insufficient guidance iterations.}
    \label{fig:perflow}
\end{figure}

Since our geometry guidance operates at each denoising step, fewer steps would proportionally reduce the number of 3DGS refitting iterations and overall runtime. However, with only 4-8 guidance opportunities, the feedback loop cannot progressively resolve cross-view inconsistencies (Fig.~\ref{fig:perflow}). While the per-view diffusion outputs appear individually plausible, the 3DGS renderings reveal clear multi-view artifacts, comparable to running our full method without guidance. A sufficient number of guidance steps is essential for our iterative framework; efficient guidance scheduling under few-step sampling remains an interesting direction for future work.

\subsubsection{Guided Forward Diffusion.}
We further experimented with a geometry-anchored forward diffusion process. The standard img2img approach adds noise to the input images in a single step to reach timestep $t_{\text{start}}$, which can introduce 3D-inconsistent structure into the noisy latents before denoising even begins. To mitigate this, we proposed a Markovian forward process where noise is added incrementally across multiple steps, and at each forward step we refit the 3DGS model to the partially-noised predictions and re-render from all views, ensuring that the noise added at each step is geometrically consistent across viewpoints. While this approach is theoretically well-motivated and did not degrade output quality, in practice it yielded only marginal improvements in 3D consistency over the standard single-step noise addition, likely because the mixed noise is already sufficient to correct any inconsistencies introduced during the forward process. The primary drawback is that this guided forward process roughly doubles the inference time, making it impractical for our pipeline. We therefore omit it from the final method.

\subsection{Full list of related work}
\label{sec:supp_related}
We now present a full list of related works. We introduce the background in \S\ref{sec:related}.
\subsubsection{Single-Image 3D Face and Head Reconstruction}
\paragraph{Parametric morphable models}~\cite{blanz2003face,smith2020morphable,blanz2023morphable,li2020learning}

\paragraph{NeRF-based}~\cite{mildenhall2021nerf,tretschk2021non,park2021nerfies,zhuang2022mofanerf,hong2022headnerf,gafni2021dynamic,buehler2024cafca,trevithick2023real}

\paragraph{3DGS-based}~\cite{kerbl20233d,xu20243d,dhamo2024headgas,wei2025graphavatar,ma20243d,qian2024gaussianavatars,shao2024splattingavatar,xiang2024flashavatar,kocabas2024hugs,xu2024gaussian,qian20243dgs,saunders2025gasp,Yan_2025_WACV,hu2024gaussianavatar,wang2025mega,giebenhain2024npga,xue2024human}. 

\paragraph{Feed-forward 3DGS}\cite{ki2024learning,li2023generalizable,ma2023otavatar,tran2024voodoo,yang2020facescape,zielonka2022towards,li2024talkinggaussian,he2025lam,zheng2023pointavatar,li2023generalizable,li2023one,chu2024generalizable,ma2024cvthead,yang2024learning,chu2024gpavatar,ye2024real3d,Wang_2025_CVPR,liang2025fastavatar}

\subsubsection{Coupling 2D Models with 3D Representations}
\paragraph{2D generative models on faces} \cite{razavi2019generating,karras2019style,karras2020analyzing,karras2021alias,dhariwal2021diffusion}

\paragraph{3D-aware training on 2D models} \cite{chan2022efficient,deng2022gram,sun2023next3d,an2023panohead,li2024spherehead,sun2022ide,hong2022headnerf,liang2023benchmarking}

\paragraph{SDS and its variants} \cite{poole2022dreamfusion,wang2023prolificdreamer,chen2023fantasia3d,liang2024luciddreamer,cao2024dreamavatar,gerogiannis2025arc2avatar,zhou2024headstudio}

\paragraph{3DGS editing} \cite{chen2024gaussianeditor,wang2024gaussianeditor,wang2024view,wen2025intergsedit,wu2024gaussctrl,chen2024dge,zhuang2024tip,lee2025editsplat,liu2024stylegaussian}

\paragraph{Image-based conditioning tokens} \cite{ye2023ip,wang2024instantid,li2024photomaker,li2023blip,wei2023elite}




\end{document}